\documentclass[twoside,11pt]{article}

\usepackage[english]{babel}
\usepackage[utf8]{inputenc}
\usepackage[T1]{fontenc}    \usepackage{amsmath}
\usepackage{graphicx}
\usepackage{subcaption}
\usepackage{cite}
\usepackage[round]{natbib}
\usepackage{dsfont}
\usepackage{url}            \usepackage{booktabs}       \usepackage{amsfonts,amssymb,bm}
\usepackage{nicefrac}       \usepackage{microtype}      
\usepackage{tikz}
\usetikzlibrary{fit,positioning}

\usepackage{jmlr2e}
\usepackage{hyperref}
\hypersetup{
	colorlinks=true,
	citecolor=red,
	filecolor=magenta, 
	urlcolor=cyan,
	linkbordercolor={1 1 1},
	citebordercolor={1 1 1}
}
\usepackage[capitalise]{cleveref}

\usepackage{subcaption} \usepackage{enumitem}
\usepackage{csquotes}
\usepackage{xfrac}
\usepackage{siunitx}
\usepackage{algpseudocode,algorithm,algorithmicx} \usepackage{wrapfig}

\graphicspath{{Pics/}}

\firstpageno{1}

\newcommand{\dom}{\mathcal{X}}
\newcommand{\domsta}{\mathcal{X}_s}
\newcommand{\domstadef}[1]{\mathcal{X}_s = \left\lbrace x : #1 \right\rbrace}
\newcommand{\domuns}{\mathcal{X}_u}

\newcommand{\inx}{x}
\newcommand{\inxopt}{x_\text{min}}
\newcommand{\fmin}{f_\text{min}}
\newcommand{\fmincons}{\tilde{f}_\text{min}}
\newcommand{\query}{\left\lbrace x,y \right\rbrace}
\newcommand{\userconf}{1-\delta}

\newcommand{\stdnoise}{\sigma}
\newcommand{\varnoise}{\stdnoise^2}
\newcommand{\thres}{c}
\newcommand{\threscons}{c_j}
\newcommand{\thresopt}{\hat{\thres}}

\newcommand{\ls}{+1}
\newcommand{\lu}{-1}
\newcommand{\fs}{\bm{f}_\text{s}}
\newcommand{\fu}{\bm{f}_\text{u}}

\newcommand{\f}{\bm{f}}
\newcommand{\Y}{\bm{y}}

\newcommand{\sobs}{\mathcal{D}}
\newcommand{\sobscons}{\sobs_{j}}
\newcommand{\GP}[2]{\mathcal{GP}\left( #1, #2\right)}

\newcommand{\Xu}{X_\text{u}}
\newcommand{\Xs}{X_\text{s}}
\newcommand{\Ys}{\Y_\text{s}}

\newcommand{\Ns}{N_\text{s}}
\newcommand{\Nu}{N_\text{u}}
\newcommand{\N}[2]{\mathcal{N}\left( #1 ; #2 \right)}
\newcommand{\mt}{\tilde{\bm{m}}}
\newcommand{\vart}{\tilde{\Sigma}}
\newcommand{\mh}{\hat{\bm{m}}}
\newcommand{\varh}{\hat{\Sigma}}
\newcommand{\F}[1]{Z_\text{EP}(#1)}
\newcommand{\E}[1]{\mathbb{E}\left[ #1 \right]}

\newcommand{\gn}{g_j}

\newcommand{\Prob}[1]{\text{Pr}\left[ #1 \right]}
\newcommand{\prob}[1]{p\left( #1 \right)}
\newcommand{\cdf}[1]{\Phi\left( #1 \right)}
\newcommand{\Id}{\bm{I}}
\newcommand{\deter}[1]{\text{det}\left( #1 \right)}
\newcommand{\ZEP}{Z_\text{EP}}
\newcommand{\df}{\text{d}\f}
\newcommand{\dfs}{\text{d}\fs}
\newcommand{\dfu}{\text{d}\fu}
\newcommand{\mesco}{{\sf mESCO}}
\newcommand{\mes}{{\sf mES}}

\newcommand{\xtrue}{x_\text{min}}
\newcommand{\xbg}{x_\text{bg}}
\newcommand{\meanbg}{\mu_\text{bg}}
\newcommand{\varbg}{\sigma^2_\text{bg}}

\newcommand{\gmaxn}{c_{j}}

\newcommand{\R}{\mathbb{R}}

\newcommand{\Mm}{M_\text{L}}
\newcommand{\Mb}{M_\text{B}}

\newcommand{\meanpost}{\bm{\mu}_\text{EP}}
\newcommand{\varpost}{\bm{\Sigma}_\text{EP}}

\newcommand{\acqui}{\alpha}

\newcommand{\acquiFailOptiC}{\tilde{\alpha}}
\newcommand{\acquimes}{\alpha_\text{\sf mES}}
\newcommand{\acquimesco}{\alpha_\text{\sf mESCO}}

\newcommand{\xnext}{x_\text{next}}

\newcommand{\gpcr}{{\sf GPCR}}

\newcommand{\MI}[2]{ \text{MI} \left( #1 ; #2 \right) }

\newcommand{\diag}[1]{\text{diag}\left( #1 \right)}

\crefname{equation}{}{} \crefname{section}{Sec.}{Sec.}
\crefname{figure}{Fig.}{Fig.}
\crefname{lemma}{Lemma}{Lemma}

\DeclareMathOperator*{\argmaxaux}{argmax}
\newcommand{\argmax}{\displaystyle\argmaxaux}
\DeclareMathOperator*{\argminaux}{argmin}
\newcommand{\argmin}{\displaystyle\argminaux}

\newcommand*\Let[2]{\State #1 $\gets$ #2}
\algrenewcommand\algorithmicrequire{\textbf{Initialize:}}
\algrenewcommand\algorithmicensure{\textbf{Postcondition:}}

\begin{document}

\title{Classified Regression for Bayesian Optimization: \\ Robot Learning with Unknown Penalties}

\author{\name Alonso Marco$^{1}$ \email amarco@tue.mpg.de \\
				\name Dominik Baumann$^{1}$ \email dbaumann@tue.mpg.de \\
	      \name Philipp Hennig$^{2}$ \email ph@tue.mpg.de \\
				\name Sebastian Trimpe$^{1}$ \email trimpe@is.mpg.de \\
				\AND
				 \addr 1. Intelligent Control Systems\\
				 Max Planck Institute for Intelligent Systems\\
	       Heisenbergstr. 3\\
	       70569 Stuttgart, Germany
		      \AND
	       \addr 2. Computer Science Department, University of T\"{u}bingen and\\
	       Max Planck Institute for Intelligent Systems\\
	       Maria von Linden Str. 6\\
	       72076 T\"{u}bingen, Germany}

\editor{ }

\maketitle

\begin{abstract}Learning robot controllers by minimizing a black-box objective cost using Bayesian optimization (BO) can be time-consuming and challenging. It is very often the case that some roll-outs result in failure behaviors, causing premature experiment detention. In such cases, the designer is forced to decide on heuristic cost penalties because the acquired data is often scarce, or not comparable with that of the stable policies. To overcome this, we propose a Bayesian model that captures exactly what we know about the cost of unstable controllers prior to data collection: Nothing, except that it should be a somewhat large number. The resulting Bayesian model, approximated with a Gaussian process, predicts high cost values in regions where failures are likely to occur. In this way, the model guides the BO exploration toward regions of stability. 
We demonstrate the benefits of the proposed model in several illustrative and statistical synthetic benchmarks, and also in experiments on a real robotic platform. In addition, we propose and experimentally validate a new BO method to account for unknown constraints. Such method is an extension of Max-Value Entropy Search, a recent information-theoretic method, to solve unconstrained global optimization problems.
\end{abstract}
 \section{Introduction}
\label{sec:introduction}

When learning a policy on a robot, it is very often the case that some roll-outs result in failure behaviors, which forbid the robot from completing the task. In those cases, the system can largely (and often quickly) diverge from the desired behavior, causing the need of a premature experiment detention, for example, by pressing an emergency button. Then, it is unclear what cost should be assigned to that failure case. Intuitively, the designer would heuristically choose a high penalty so that similar policies are never visited again. 
For example,~\citep{marco16tuning} propose a method to automatically tune controller parameters of a humanoid robot that learns to balance an inverted pole. Therein, unsuccessful balancing experiments were penalized with a large number, chosen heuristically a priori. 
In~\citep{rai2017bayesian}, a walking controller of a bipedal robot is learned from data, using a tailored cost function that penalizes with a high cost roll-outs in which the robot falls down.
The authors~\citep{tu18a} learn the parameters of an LQR feedback policy from data in a reinforcement learning (RL) setting, and assign an infinite cost when the policy destabilizes the closed-loop behavior.
Finally, in~\citep{kober2009motor}, RL is used for the ball-in-a-cup problem, assigning a constant cost to roll-outs in which the ball does not pass upwards the rim of the cup.

While these methods show successful
learning experiments, choosing appropriate penalties for failure cases is a key component of the reward design, and unclear a priori~\citep{chatzilygeroudis2018survey}. 
In failure cases (e.g., unstable controllers), deciding on a penalty is non-trivial, because:
\begin{enumerate}[label=(\alph*)]
\item The immediate cost values might be orders of magnitude too large and not comparable against those of the stable policies.
\item Due to the premature experiment detention, the collected data can be too scarce, and not comparable with that of the stable policies, whose experiments lasted for longer time~\citep{marco16tuning}.
\item After stopping the experiment, sensor measurements could be difficult, or impossible to access.
\end{enumerate}

\noindent To circumvent these problems, the cost designer typically ignores the acquired data at failure cases and commits to \emph{ad hoc} choices, which are decided and fixed before doing any experiment. While this might sound reasonable, \emph{prior to any data collection} and without expert knowledge about the system, the designer does not know how to choose such penalty. Furthermore, arbitrary choices come at the risk of laborious and repeated manual tuning:
During learning it is possible that stable roll-outs retrieve a cost worse than the arbitrary penalty chosen a priori, which forces a re-design of the cost function, and thus, possibly a restart of the learning experiments.
Generally speaking, choosing an appropriate penalty for failures is problem-dependent and non-trivial. Moreover, failure cases need to be treated separately when designing the cost function, typically requiring manual tuning effort.

In the context of robot learning, Bayesian optimization (BO)~\citep{shahriari2016taking}
can be seen as a form of RL for policy optimization. Therein, the latent cost function represents an unknown mapping from the parameters of a policy executed on the robot to its performance on completing some task~\citep{vonrohr2018gait,rai2017bayesian,marco16tuning,calandra2016bayesian,Berkenkamp2016SafeOpt}. Such cost is typically modeled in a probabilistic setting and minimized iteratively through experiments. BO incorporates a probabilistic prior about the objective cost typically through a Gaussian process (GP) regression model \citep{Rasmussen2006Gaussian}. This is a non-parametric model that allows for inference on the latent cost at unobserved locations, conditioned on collected data. 

The core contribution of this paper is a novel Bayesian model for the cost function, such that the penalty term does not need to be defined a priori, but instead is learned from data.
The likelihood of the Bayesian model that can tackle a \emph{hybrid} dataset, inside which stable policies yield a continous-valued cost observation, and unstable policies only yield the discrete label \enquote{unstable}. The consequent intractable posterior is approximated to obtain a tractable GP model.
In addition, the GP model is used within BO to reveal the optimal policy by sequentially selecting experiments that avoid unstable areas with high probability. 
Because the resulting model serves both, as a probabilistic regressor, and classifies regions as stable or unstable, we call it Gaussian process for \emph{classified regression} (\gpcr).
We illustrate the problem of learning with failures with a case example.
\begin{example}
Consider tuning the policy parameters of a humanoid robot doing a simple task, like squatting. The goal is to find the parameters that minimize the tracking error of the movement. Such error is measured through a cost function that is evaluated after each experiment. During learning, we observe that some controllers are making the robot fall down, and some others cause huge vibrations on the joints. Since those cases are undesired, and considered as unstable, an operator prematurely stops the experiment pressing an emergency button. We consider all these cases as failures, for which no meaningful data is available.
\label{ex:hermes}
\end{example}

\noindent In~\cref{ex:hermes}, the unstable cases can be seen as consequence of two external constraints being violated, i.e., the velocity of the joints and the tilting angle of the torso surpassing certain thresholds. In case such thresholds are known, one can know during learning how close is a specific controller from being unstable by measuring the sensor signals.
However, prior to \emph{any} data collection, the designer does not know what those thresholds are, i.e., neither the largest joint velocities right before vibrations start to happen, nor the maximum angular position of the torso before the robot falls down.
The proposed \gpcr~model can also be used to model such constraints when the constraint thresholds are unknown. Furthermore, \gpcr~estimates such unknown thresholds from data. This is the second core contribution of the paper.

In addition, the resulting constrained problem is addressed using Bayesian optimization methods under the presence of unknown constraints (BOC). To this end, recent BOC methods motivated by information theory \citep{lobato2016general} have demonstrated to outperform existing heuristics. However, they are computationally demanding. Instead, we have extended a recent information-theoretic Bayesian optimizer, Max-Value Entropy Search~\citep{wang2017mes}, which is computationally cheaper, to account for unknown constraints. The resulting algorithm is called Min-Value Entropy Search with unknown level-set constraints (\mesco). That is the third contribution of this work.

The paper is structured as follows: 
In~\cref{sec:prob_for}, we formulate the constrained controller tuning problem, and define the key mathematical elements of the paper. In \cref{sec:ClassiReg}, we present the novel Bayesian model (\gpcr) in the context of unknown failure penalties. In \cref{sec:applications}, a range of applications of the \gpcr~model are presented, including the extension to model black-box constraints, and specially the case in which constraint thresholds are unknown. In \cref{sec:mesco}, Bayesian optimization with unknown constraints is explained, along with our extension (\mesco) of Max-Value Entropy Search. In \cref{sec:relWork}, we discuss our contributions among the existing literature. In \cref{sec:exper}, we demonstrate the benefits on \gpcr~on several numerical simulations and experiments on a real robotic platform, and we conclude with a discussion in~\cref{sec:conc}.

 \section{Preliminaries}
\label{sec:prob_for}

In this section, we start with a general formulation of the controller tuning problem. We then define mathematically the unknown instability threshold, a key parameter of the \gpcr~model, and finally introduce notation for Gaussian process (GP).

 \subsection{Controller tuning problem}
\label{ssec:controlprob}
Let a system be defined by discrete unknown dynamics $s_{t+1} = h(s_t,u_t,w_t)$, where $s_t \in \mathbb{R}^W$ is the state, $u_t \in \mathbb{R}^V$ is the control signal, and $w_t \in \mathbb{R}^W$ is process noise, at time $t$. We assume the control signal is obtained from a feedback policy $u_t = \pi(s_t)$. Generally, we are interested in policies $\pi$ that make the system accomplish a specific user-defined task, for example, picking an object with a robot arm and placing it somewhere else on the table. Such policies drive the system from the initial state $s_0$ to a terminal state $s_T$ describing a time trajectory as a sequence of states and control inputs $\tau_T = \left\lbrace s_t,u_t \right\rbrace_{t=0}^T$. The cost of the system for not accomplishing the desired
task is specified via a cost function $l: \mathbb{R}^W\times \mathbb{R}^V \rightarrow \mathbb{R}_{\geq 0}$. The cost of a full trajectory under the policy $\pi$ is then defined
as $f_\tau(\pi) = \mathbb{E}\left[\sum_{t=0}^{T-1} l(s_t,u_t)\right]$. While the true systems dynamics $h$ are unknown, we assume to have access to an approximate model $\hat{h}$. In cases in which the model $\hat{h}$ and the cost $l$ have certain structure,
it is possible to compute an optimal policy
\begin{equation}
\begin{array}{rcl}
\hat{\pi} = & \argmin_{\pi} & \mathbb{E}\left[\sum_{t=0}^{T-1} l(x_t,\pi(x_t))\right] \\
& \text{s.t.} & s_{t+1} = \hat{h}(s_t,u_t,w_t).
\end{array}
\label{eq:pol_opti}
\end{equation}
For example, when the model $\hat{h}$ is linear and the cost $l$ is quadratic, the optimal policy $\hat{\pi}$ is the linear quadratic regulator (LQR) \citep{anderson2007optimal}. Generally speaking, while such policy $\hat{\pi}$ is optimal for the model $\hat{h}$, it is suboptimal for the true dynamics $h$. In order to recover the optimal policy for $h$, we parametrize $\hat{\pi}(s_t) = \hat{\pi}(s_t; x)$ with parameters $x \in\dom \subset \mathbb{R}^D$ and learn the optimal parameters by collecting experimental data from the true system $h$. To this end, we obtain a different cost of a full trajectory $f_\tau(\tau_T(x))$ 
for each parametrization $x$.
This induces an objective $f_\tau(\tau_T(x)):\mathcal{X} \rightarrow \mathbb{R}$, which we write as $f(x)$ to simplify notation. Then, the learning problem is to find the parametrization $\inxopt$ that minimizes the trajectory cost
\begin{equation}
\inxopt = \argmin_{x\in\dom} f(x) \\
\label{eq:min_cost}
\end{equation}
on the true dynamics $s_{t+1} = h(s_t,u_t,w_t)$. The function $f$ is unknown, as it depends on unknown dynamics $h$, but we can query it as a black-box function at location $x$ by doing an experiment on the real system. For instance, in \citep{marco16tuning} the weighting matrices of an LQR controller design are parametrized with parameters $x$. Therein, the goal is to minimize an unknown cost objective $f$ that quantifies the performance of a real robot balancing an inverted pole. Thus, querying the function $f$ at location $x$ consists of doing an experiment with policy $\hat{\pi}(s_t; x)$ and collecting sensor data to compute the corresponding cost value $f(x)$.

\subsection{Instability threshold as a level-set constraint}
\label{ssec:thres}
In the RL literature, the data collected with an unstable controller is typically discarded, and a relatively large value is heuristically chosen to characterize its performance. Whereas we propose to also discard such data, we never choose any heuristic penalty. These ideas can be mathematically reflected in the objective $f$ as follows.

\noindent Let us divide the domain $\dom$ in two sets: The set of stable controllers $\domsta$, for which function values $f(x)$ are known, and the set of unstable controllers $\domuns$, for which function values $f(x)$ are somewhat large, but unknown, with $\domsta \bigcup \domuns = \dom$. To include these concepts in the objective $f$, we re-define it so that its image is
\begin{equation}
\begin{split}
f(\domuns) & \in (c,+\infty)  \\
f(\domsta) & \in [\fmin,c],
\end{split}
\label{eq:fundef}
\end{equation}
where $f$ is lower bounded by the global minimum $\fmin = \min_{x\in\domsta}f(x)$, and the image of the objective over the stable set is upper bounded by 
 a threshold $\thres$, which is defined next.

\begin{definition}
The instability threshold $\thres = \sup_{x\in\domsta} f(x)$ is the worst possible cost excluding those of the unstable set $\domuns$.
\label{def:thres}
\end{definition}

\noindent In the context of~\cref{ex:hermes}, the instability threshold $c$ can be seen as the worst squatting possible to observe without destabilizing the robot. 
From \cref{eq:fundef}, it follows that the threshold $c$ induces
a level set $\mathcal{X}_c = \left\lbrace x : f(x) = c \right\rbrace$, with $\mathcal{X}_c \subset \domsta$. In addition, we say that $\thres$ imposes a \emph{level set-constraint} in $f$ that divides the input space $\dom$ into the stable set $\domsta = \left\lbrace x : f(x) \leq \thres \right\rbrace$ and the unstable set $\domuns = \left\lbrace x : f(x) > \thres \right\rbrace$. Finally, since the values of $f(x)$ that correspond to stable controllers are ``constrained'' to live below $c$, we say that the function $f$ is \emph{self-constrained}.

It is important to remark that without complete knowledge of the function $f$, and prior to any data collection (i.e., never having realized an experiment on the system), $\thres$ is unknown to the designer. However, in~\cref{ssec:threshold} we show that $\thres$ can be estimated from data, and that such estimation is useful to target stable regions when searching for the optimum \cref{eq:min_cost}.\\

\noindent We have introduced the instability threshold $\thres$, a key concept that we use to characterize instability in the context of the image of the objective $f$. This concept is important, as it will be recurrently used in the remainder of the paper.

\subsection{Gaussian process (GP)}
\label{ssec:GP}
The Bayesian model for the objective function $f$, proposed in~\cref{sec:ClassiReg}, is approximated as a GP~\citep{Rasmussen2006Gaussian}.
This is a probability distribution over the function $f$, such that every finite restriction to random function values $\left[f(x_1),\dots,f(x_N)\right]$ follows a multivariate Gaussian distribution. 
That means, the objective function $f$ is modeled as $f \sim \GP{m(x)}{k(x,x^\prime)}$, with prior mean $m:\dom \rightarrow \mathbb{R}$, and positive definite covariance function (also called kernel) $k:\dom \times \dom \rightarrow \mathbb{R}$. The kernel encodes the strength of statistical correlation between two latent function values $f(x_i)$ and $f(x_j)$, where $x_i,x_j \in \dom $. The model proposed in~\cref{sec:ClassiReg} assumes a zero-mean prior, $m(x) \equiv 0$.\\\\
\noindent
In the remainder of this paper we will refer to \emph{controller parametrization} $x$ simply as \emph{controller} $x$. In addition, we will refer to a \emph{stable/unstable controller} $x$ as stable/unstable $x$.

 \section{Gaussian process model for classified regression (\gpcr)}
\label{sec:ClassiReg}
In the following, we present the Bayesian model for classified regression and how to approximate it as a Gaussian process. Such model for the objective function $f$ is presented in~\cref{eq:min_cost}. This model inherently encodes the particular distinction between unstable and stable controllers through the instability threshold $\thres$ defined in~\cref{ssec:thres}.

\subsection{Likelihood model for unknown penalties}
\label{ssec:lik}
Let us assume that for each parameter $\inx_i$ we have access to two types of observations: (i) a binary label $l_i\in \left\lbrace \lu,\ls\right\rbrace$ that determines whether the experiment was unstable or stable, and (ii) a noisy cost value $y_i \in \R$ only when the experiment is stable:
\begin{equation}
\label{eq:output}
(y_i,l_i) = 
\left\lbrace
\begin{array}{llr}
(f_i + \varepsilon, &\ls), & \text{ if } x\in\domsta \\
(\varnothing,&\lu), & \text{ if } x\in\domuns \\
\end{array}
\right.
,
\end{equation}
where we have used the shorthand notation $f_i = f(\inx_i)$, $y_i = y(\inx_i)$, and $l_i = l(x_i)$.
As stated in~\cref{ssec:thres},
we assume that all stable controller parametrizations $\inx_i\in \domsta$ yield noise-corrupted observations $y_i = f_i + \varepsilon$, with $\varepsilon \sim \N{\varepsilon}{0,\varnoise}$.
In the context of the problem described in~\cref{ssec:controlprob}, the observation noise arises from
 the process noise $w_t$ in the system, and the need of approximating the expectation from~\cref{eq:pol_opti} in practice \citep{marco16tuning}.
For unstable controller parametrizations $\inx_i\in \domuns$, we assume that no continuous-valued observation was obtained, but only a discrete label.

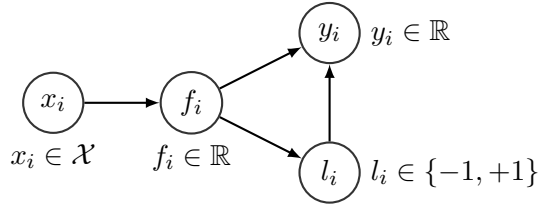
\begin{figure}[t]
\centering
\begin{tikzpicture}
\tikzstyle{main}=[circle, minimum size = 8mm, thick, draw =black!80, node distance = 1mm and 10mm]
\tikzstyle{connect}=[-latex, thick]
\tikzstyle{box}=[rectangle, draw=black!100]
  \node[main, fill = white!100] (x) [label=below:$\inx_i \in \dom$] {$\inx_i$};
  \node[main] (f) [right=of x,label=below:$f_i \in \mathbb{R}$] {$f_i$};
  \node[main, transparent] (w) [right=of f] { };
    \node[main] (y) [above=of w,label=right:{$y_i\in\R$}] {$y_i$};
  \node[main] (l) [below=of w,label=right:{$l_i\in\left\lbrace-1,+1\right\rbrace$}] {$l_i$};
  \path (x) edge [connect] (f)
        (f) edge [connect] (y)
        (f) edge [connect] (l)
        (l) edge [connect] (y);
\end{tikzpicture}
\caption{Graphical model for Classified Regression\protect\footnotemark}
\label{fig:model}
\end{figure}

The graphical model in~\cref{fig:model} expresses the relation between variables. For a given $\inx_i$, having full information about $f_i$ is sufficient to determine whether the controller is stable or not ($l_i$), and whether we have access to a cost value observation or not ($y_i$). Notice that the noisy evaluation $y_i$ is not fully determined by $f_i$, but also requires the additional information about the associated label $l_i$ (cf.\ \eqref{eq:output}). Such model can be encoded in
a likelihood over the observation $\left( y_i,l_i\right)$ of the latent function $f_i$ at location $x_i$ as
\begin{equation}
p(y_i,l_i|f_i,x_i) = 
H(l_i(\thres-f_i))\left(1_{\{l_i = \ls\}}\N{y_i}{f_i,\varnoise} + 1_{\{l_i = \lu\}} \right),
\label{eq:lik}
\end{equation}
where the dependency on $x_i$ in the right-hand side is redundant, but it has been kept for clarity.
$H(z)$ is the Heaviside function, i.e., $H(z) = 1$, if $z \geq 0$, and $H(z) = 0$ otherwise. The likelihood \cref{eq:lik} captures our knowledge about the latent function: When $x_i$ is unstable, all we know about $f_i$ is that it takes any possible value above the threshold $\thres$, with all values $f_i\geq\thres$ being equally likely, but we never specify what that value is. If $x_i$ is stable, all the probability mass falls below $\thres$ for consistency, and the indicator function $1_{\{ \cdot \}}$ toggles the contribution of the Gaussian.

Let the latent objective $f$ be observed at locations $X=\{X_\text{s},X_\text{u}\}$, which entails both, stable $X_\text{s}=\left\lbrace x_i\right\rbrace_{i=1}^{\Ns}$ and unstable controllers $X_\text{u}=\left\lbrace x_i\right\rbrace_{i=\Ns+1}^N$, with $N = \Ns+\Nu$. The corresponding latent objective values at the stable locations $X_\text{s}$ are $\fs=\left[f_1,f_2,\dots,f_{\Ns}\right]^\top$, and $\fu=\left[f_{\Ns+1},f_{\Ns+2},\dots,f_N\right]^\top$ at $X_\text{u}$. We group the latent objective values in a vector $\f = \left[ \fs^\top, \fu^\top \right]^\top$.
The latent function values induce observations grouped in the set $\sobs = \left\lbrace y_i,l_i \right\rbrace_{i=1}^{\Ns} \cup \left\lbrace l_i \right\rbrace_{i={\Ns+1}}^N$. Since $\sobs$ contains both discrete labels and real scalar values, we refer to it as a \emph{hybrid} set of observations. 
We assume such observations to be conditionally independent and identically distributed following~\cref{eq:lik}, which allows the likelihood to factorize $p(\sobs|\f,X) = \prod_{i=1}^N p(y_i,l_i|f_i,x_i)$. Then, the likelihood over the data set $\sobs$ becomes
\begin{equation}
p(\sobs|\f,X) = \prod_{i=1}^{\Ns} H(\thres-f_i) \mathcal{N}(y_i|f_i,\varnoise) \hspace{-7pt}  \prod_{i=\Ns+1}^N \hspace{-7pt}  H(f_i-\thres).
\label{eq:lik_elabo}
\end{equation}

\footnotetext{While the objective function \cref{eq:fundef} is lower bounded by $\fmin$, an instance $f_i$ at location $x_i$ has infinite support. While this might seem contradictory, it is actually possible because \cref{eq:fundef} represents the true (unknown) objective, and $f_i$,$y_i$ are random variables, assumed to be Gaussian distributed, and thus $f_i,y_i \in \R$.}

\subsection{Posterior probability}
\label{ssec:posterior}
The posterior follows from Bayes theorem
\begin{equation}
p(\f|\sobs,X) \propto p(\sobs|\f,X)p(\f|X),
\label{eq:bayes}
\end{equation}
with zero-mean Gaussian prior $p(\f|X) = \mathcal{N}(\f|\bm{0},\bm{K})$
and likelihood~\cref{eq:lik_elabo}.
The last term in~\cref{eq:lik_elabo} can be rewritten as a multivariate Gaussian $\prod_{i=1}^{\Ns} \mathcal{N}(y_i|f_i,\varnoise)=\mathcal{N}(\Ys|\fs,\varnoise \bm{I})$.
Then, we obtain the posterior as
\begin{equation}
p(\f|\sobs,X) \propto 
\prod_{i=1}^{\Ns} H(\thres-f_i) \hspace{-7pt}  \prod_{i=\Ns+1}^N \hspace{-7pt}  H(f_i-\thres)
\mathcal{N}(\Ys|\fs,\varnoise \bm{I})
\mathcal{N}
\left(
\begin{bmatrix} \fs \\ \fu \end{bmatrix}
\Big\vert
\begin{bmatrix} \bm{0} \\ \bm{0} \end{bmatrix}
,
\begin{bmatrix} K_{ss} & K_{su} \\ K_{us} & K_{uu} \\ \end{bmatrix}
\right),
\label{eq:post_elabo}
\end{equation}
where $K_{ss}$, $K_{su} = K_{us}$, and $K_{uu}$ are the prior covariance matrices expressing the correlation between function evaluations at stable (s) and/or unstable (u) locations. The last two terms are the product of two multivariate Gaussians of different dimensionality. This is equal to another unnormalized Gaussian $\alpha\N{\f}{\tilde{\bm{m}},\tilde{\bm{\Sigma}}}$, whose mean, variance and scaling factor depend on the observations and the noise, i.e.,
$\tilde{\bm{m}} = \tilde{\bm{m}}(\Ys,\varnoise)$,
$\tilde{\bm{\Sigma}} = \tilde{\bm{\Sigma}}(\varnoise)$
and $\alpha = \alpha(\Ys,\varnoise)$, respectively (cf. \cref{app:prodGauss}).
Hence, the posterior becomes
\begin{equation}
p(\f|\sobs,X) \propto \prod_{i=1}^{\Ns} H(\thres-f_i) \hspace{-7pt}  \prod_{i=\Ns+1}^N \hspace{-7pt}  H(f_i-\thres) \N{\f}{\mt,\vart},
\label{eq:my_post}
\end{equation}
where the scaling factor $\alpha$ has been omitted for simplicity. The role of the Heaviside functions is to restrict the support of $p(\f|\sobs,X)$ to an unbounded hyperrectangle with a single corner at location $f_i=c,\;\forall i \in \{1,\dots,N\}$. Thus, \cref{eq:my_post} can be geometrically understood as the product of a multivariate Gaussian density of dimension $N$ times an activation function that sets to zero any probability outside such hyperrectangle. Computing the right-hand side of \cref{eq:my_post} is analytically intractable. However, it can be approximated with an unnormalized Gaussian 
\begin{equation}
p(\f|\sobs,X) \propto q(\f|\sobs,X) = \ZEP\N{\f}{\meanpost,\varpost},
\label{eq:posterior_approx}
\end{equation}
where $\ZEP$, $\meanpost$, and $\varpost$ are computed following~\citep{cunningham2011gaussian}. Therein, the posterior of a multivariate Gaussian, whose support is restricted to bounded polyhedra (i.e., a more general case than the one from~\cref{eq:my_post}), is approximated using Expectation propagation (EP)~\citep{minka2001expectation}.
Our implementation of such EP approximation performs comparably to computing the posterior using sampling methods, e.g., Elliptical Slice Sampling~\citep{murray2010slice}. However, we have found EP to need much lower wall-clock time to yield the same results.

\subsection{Predictive probability}
\label{ssec:predictive}

Our goal is to make inference at an unobserved location $x_*$, where the predictive latent function value $f_* = f(x_*)$ is jointly distributed with the vector of latent function values at the observations $\f$, i.e., $p(\f,f_*)$. The joint posterior is
\begin{equation}
p(\f,f_*|\sobs,X,x_*) \propto p(\sobs|\f,X) p(\f,f_*|X,x_*),
\label{eq:prior_extended}
\end{equation}
where the likelihood term does not depend on the unobserved location $x_*$, nor on $f_*$.
The predictive distribution of $f_*$ conditioned on the collected data can then be obtained by marginalizing over the latent function values $\f$ from the joint posterior as
\begin{equation}
p(f_*|\sobs,X,x_*) \propto \int_{\f} p(\sobs|\f,X) p(\f,f_*|X,x_*)\text{d}\f,
\label{eq:pred}
\end{equation}
where the likelihood term is defined as in~\cref{eq:lik_elabo}, and we extend the prior with the unobserved location $\inx_*$ as
\begin{equation}
p(\f,f_*|X,x_*)
=
\N{\begin{bmatrix} \f \\ f_* \end{bmatrix}}
{\begin{bmatrix} \bm{0} \\ 0 \end{bmatrix}, \begin{bmatrix} K_{\f\f} & K_{\f *}  \\ K_{*\f} & K_{**} \end{bmatrix}}
.
\label{eq:prior}
\end{equation}
The multivariate integral in~\cref{eq:pred} is analytically intractable due to the structure of the likelihood \cref{eq:lik_elabo}. 
However, we can alleviate this by applying some transformations. First, by combining  $p(\f,f_*|X,x_*) = p(\f|X)p(f_* | \f,X,x_*)$ and \cref{eq:bayes}, the integral from \cref{eq:pred} can be rewritten as $\int_{\f} p(\f|\sobs,X)p(f_* | \f,X,x_*)\text{d}\f$. Then, using the Gaussian approximation \cref{eq:posterior_approx} of the posterior $p(\f|\sobs,X)$, the resulting predictive distribution is also Gaussian $q(f_*)=\N{f_*}{\mu_*,\sigma^2_*}$, with mean $\mu_*$ and variance $\sigma^2_*$ as a function of the mean $\meanpost$ and covariance $\varpost$ of the posterior~\cite[Sec. 3.4.2]{Rasmussen2006Gaussian}. Using such approximated posterior~\cref{eq:posterior_approx}, the moments of $q(f_*)$ are analytically given by
\begin{equation}
\begin{array}{rl}
\mu_* = & K_{*\f}K_{\f\f}^{-1}\meanpost \\ 
\sigma^2_* = & K_{**} - K_{*\f}K_{\f\f}^{-1} \left( \bm{I} +  \varpost K_{\f\f}^{-1} \right) K_{\f*}.
\end{array}
\label{eq:pred_q_moments}
\end{equation}
We illustrate in~\cref{fig:GPCR_cons} (top) the approximate posterior $q(f_*)$ for the setting described in \cref{ex:onedimGP}. A detailed analysis of the quality of the approximation $q(f_*)$ to $p(f_*|\sobs,X,x_*)$ can be found in~\cref{app:exact_pred}. 
\begin{example}
Consider the unknown objective $f:\mathcal{X} \rightarrow \mathbb{R}$, $\mathcal{X} = \left[0,1\right]$, for which we have three stable evaluations $\Xs=\left[0.1,0.3,0.5\right]$, $\Ys=\left[0.5,2.0,1.0\right]$ and two unstable evaluations $\Xu=\left[0.7,0.9\right]$. We assume small noise $\sigma_n = 0.02$, and a zero-mean prior with covariance $\bm{K}$, whose entries $K_{ij} = k(x_i,x_j)$ are computed with an ARD Mat\'{e}rn $3/2$ kernel $k$ with signal variance $0.5$ and lengthscale $0.2$.
\label{ex:onedimGP}
\end{example}

\noindent
The univariate distribution $q(f_*)$ is straightforwardly extended to a set of unobserved locations $\bm{x}_*$, where we consider the joint prior $p(\f,\f_*)$. This gives rise to a multivariate approximate Gaussian posterior $q(\f_*) = \N{\f_*}{\bm{\mu}_*,\bm{\sigma}_*^2}$, i.e., a Gaussian process.

The threshold $\thres$ influences the predictive distribution through the likelihood model~\cref{eq:lik_elabo}, which induces an intractable posterior that, when approximated as a multivariate Gaussian, yields a Gaussian process model. In addition, the threshold $\thres$ can be seen as a discriminator that distinguishes instability from stability in the axis of the cost value. Because of the closeness to the idea of classification, but reinterpreted in the regressor, we call this model: Gaussian process for \emph{classified regression} (\gpcr).\\

\noindent In the next section we give detailed insights about the influence of $\thres$ in the model, and also how it can be learned from data.

\subsection{Instability threshold as a model parameter}
\label{ssec:threshold}
The instability threshold $\thres$ plays an important role in the likelihood model~\cref{eq:lik_elabo}.
However, prior to data observation, one has in principle no correct notion about an appropriate value for it.
Instead, we can compute an estimate $\thresopt$ from data using \gpcr. Herein, we show two possible ways of computing $\thresopt$.

\subsubsection{Estimation via maximum likelihood (ML)}
One possible way to estimate $\thres$ from data is by maximizing the marginal likelihood $p(\sobs|X)$, which can be computed by integrating out $\f$ from the right-hand side of~\cref{eq:my_post}. Since that is analytically intractable, we instead use the integral over the approximate posterior $q(\f|\sobs,X)$ \cref{eq:posterior_approx} to compute its marginal
\begin{equation}
\thresopt = \argmax_{\thres}\; \log\int_{\f}q(\f|\sobs,X,\thres)\df,
\label{eq:c_opti}
\end{equation}
where the dependency on $\thres$ has been introduced in $q(\f|\sobs,X)$ for clarity. From~\cref{eq:posterior_approx} follows that $\int_{\f}q(\f|\sobs,X,\thres)\df = \F{\thres}$. Thus, solving~\cref{eq:c_opti} implies performing a line search, for which every call to the function $\F{\thres}$ involves solving an EP loop.

\subsubsection{Estimation via maximum a posteriori (MAP)}
\label{sssec:MAP}
Another possible way of estimating $\thres$ is to consider it as a stochastic variable distributed with a \emph{hyperprior} $p(\thres)$, and 
compute the maximum of the posterior distribution $p(\thres|\sobs) =  \int_{\f}p(\sobs|\f,X,\thres)p(\f|X)p(\thres)\df$, where we have assumed that $\thres$ and $\f$ are uncorrelated. Using the approximate posterior~\cref{eq:posterior_approx} over $\f$, and Bayes' rule, we 
obtain the posterior
$
p(\thres|\sobs) \propto \int_{\f} q(\f|\sobs,X,\thres) p(\thres)\df.
\label{eq:post_c}
$
Due to the lack of prior knowledge about $\thres$, we propose a wide Gaussian hyperprior $p(\thres)=\N{c}{\mu_c,\sigma_c^2}$, whose parameters are to be selected by the designer. Similar to~\cref{eq:c_opti}, we find $\thresopt$ by solving the optimization problem

\begin{equation}
\thresopt = \argmax_{\thres}\; \log\F{\thres} - 0.5(c-\mu_c)^2/\sigma_c^2.
\label{eq:c_opti_map}
\end{equation}

An advantage of ML over MAP is that the former requires no prior reasoning about $\thres$ at all. However, we have found in our experiments that ML can lead to inconvenient estimations of $c$ when stable evaluations are absent. In such case, we see from~\cref{eq:my_post} that the posterior probability becomes larger since the term $\prod_{i=\Ns+1}^N H(f_i-\thres)$ activates a larger portion of the Gaussian, and has a maximum at $\thresopt=-\infty$. Such estimation can cause problems when using the model to perform Bayesian optimization under unknown constraints, as explained in~\cref{sec:mesco}. Analogously, when only stable evaluations are present, the maximum is found at $\thresopt=+\infty$. On the contrary, MAP enjoys the hyperprior over $\thres$ as a regularizer that avoids such extreme point estimates.

The estimated instability threshold $\thresopt$ becomes important when determining the probability of a specific unforeseen location $x$ resulting in a stable controller. In the next section, we describe how to compute such probability.

\subsection{Probabilistic stability constraint in \gpcr}
\label{ssec:probcons}

Given our current estimate of the instability threshold $\thresopt$, we can compute the probability of a specific point $x$ being a stable observation, i.e., $\Prob{x \in \domsta}$. From~\cref{ssec:thres}, it follows that $\Prob{x \in \domsta} = \Prob{f(x) \leq \thresopt}$. 
After having observed data $\sobs$, we define
\begin{equation}
\Prob{f(x) \leq \thresopt \mid \sobs} = \int_{-\infty}^{\thresopt}\prob{f|\sobs,x} \text{d}f \simeq \int_{-\infty}^{\thresopt}q \left( f \right) \text{d}f = \cdf{z(x)},
\label{eq:predictive_constraint}
\end{equation}
where the predictive distribution \cref{eq:pred} has been rewritten as $\prob{f|\sobs,x}$ for simplicity, and its approximation~\cref{eq:pred_q_moments} as $q(f) = \N{f}{\mu(x),\sigma^2(x)}$, where $\Phi$ is the CDF of a univariate Gaussian distribution, and $z(x) = (\thresopt-\mu(x))/\sigma(x)$.
\begin{figure}[t!]
\centering
\includegraphics[width=0.8\columnwidth]{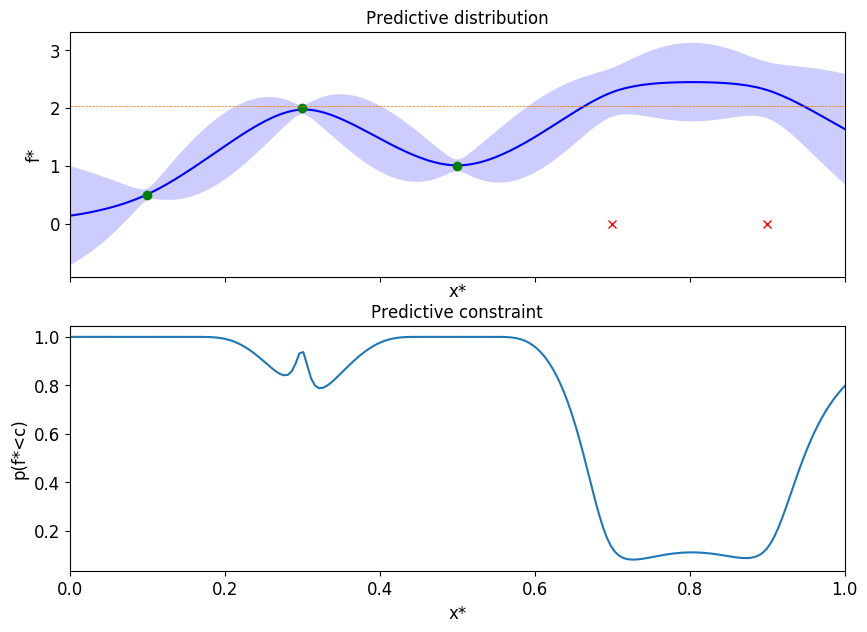}
\caption{(top) \gpcr~model posterior conditioned on three stable evaluations (green dots), and two unstable evaluations (red crosses). The estimated instability threshold $\thresopt$ (orange dashed line) is used by the \gpcr~model to push up the probability mass in those regions where unstable observations are predicted. (bottom) Probabilistic constraint
$\Prob{f(x) < \thresopt \mid \sobs}$
of the \gpcr~model, as detailed in~\cref{eq:predictive_constraint}. Both plots follow the settings from \cref{ex:onedimGP}.}
\label{fig:GPCR_cons}
\end{figure}
The predictive constraint~\cref{eq:predictive_constraint} is depicted in~\cref{fig:GPCR_cons}, using the synthetic~\cref{ex:onedimGP}.

 \section{Applications and extensions of \gpcr}
\label{sec:applications}
The \gpcr~framework introduced in~\cref{sec:ClassiReg} can be used to model the objective function from~\cref{eq:min_cost}, but it is not limited to it. One could think about the class of unstable controllers as something wider, i.e., any controller that results in any undesired behavior. For example, a robot arm achieving a stable trajectory, but hitting an obstacle on the way. Such undesired behavior can be measured, for example, by computing the euclidean distance from the robot arm to the obstacle. 
This information is much richer than a discrete label to indicate success/failure and could, for instance, be treated as an additional external black-box function, modeled as a standard GP
\footnote{By ``standard GP'' we mean the most common use of a GP, i.e., using a Gaussian likelihood, as opposed to the \gpcr~model proposed herein.}
, and same for the objective. In such a case, at first sight it seems like Bayesian optimization under unknown constraints \citep{gelbart2014bayesian,gardner2014bayesian} suffices to address the constrained problem and \gpcr~is not needed. However,
there exist many scenarios in which \gpcr~comes at handy; for example, when the constraint threshold is unknown (i.e., in the aforementioned example with the robot arm, we do not know a priori where the obstacle is). In that case, one can model the constraint using \gpcr, and see the unknown constraint threshold as the instability threshold introduced in \cref{ssec:threshold}. By doing so, we could also learn the constraint threshold from data.

\noindent Including the aforementioned case, we have identified in total four different cases where the \gpcr~model plays an important role in either modeling the constraint, the objective, or both.
However, before explaining each case, we need to introduce some new notation and concepts that will make the differences clear. Let us assume the optimization problem~\cref{eq:min_cost} is constrained under $M \geq 1$ black-box external constraints, which results in the 
constrained minimization problem
\begin{equation}
\begin{aligned}
\fmincons = & \min_{x\in\domsta} f(x) \\
& \text{with  } \domsta := \left\lbrace x : g_1(x) \leq c_1,\cdots,g_M(x) \leq c_M \right\rbrace,
\end{aligned}
\label{eq:min_cost_cons}
\end{equation}
where $c_j \in \mathbb{R},\;\forall j=\{1,\cdots,M\}$ are the constraint thresholds. 
From now on, we assume that evaluations are \emph{coupled}, i.e., when evaluating at $x_i$, we obtain measurements from both the objective, and all the constraints \citep{lobato2016general}. Additionally, we distinguish between ``stable'' and ``safe'' controllers: A controller $x_i$ is considered unsafe, but stable, when it violates at least one of the constraints, but it keeps the system stable (e.g., the robot arm performs a stable trajectory, but it finds an obstacle in the way). Similarly, $x_i$ is considered safe, but unstable, when it does not violate any constraint, but destabilizes the system (e.g., the robot arm executes a poor controller, which destabilizes the system, but it does not hit any obstacle). Consequently, we redefine $\domsta$ more broadly as to include both, stable and safe controllers, as opposed to the definition from \cref{ssec:thres}. Although the evaluations are coupled ``in location'', we assume they are uncoupled ``in observation''. For example, an unsafe but stable controller yields a discrete label when evaluating the constraint, but a continuous-valued observation when evaluating the objective.

The nature of the constraints \cref{eq:min_cost_cons} is important when deciding whether the \gpcr~model is useful or not. More concretely, when it comes to obtain a query from a constraint $g_j$, we distinguish two characteristic cases: (i) the query value $g_j(x)$ exists when it is satisfied, and it does not exist when it is violated (instead a binary label is obtained), and (ii) the query value $g_j(x)$ does not exist in either case, and we only obtain a binary outcome that determines constraint satisfaction/violation. In the first case, we characterize $g_j$ as a \emph{level-set constraint}, while in the second case, we characterize it as a \emph{binary constraint}. For modeling a level-set constraint, \gpcr~has benefits over standard GPs, because it can estimate the constraint threshold during learning. For modeling a \emph{binary constraint}, \gpcr~has no special benefit, as the dataset is not hybrid (cf. \cref{ssec:lik}), but binary.
Given these two clearly distinctive groups, let us assume $\Mm$ level-set constraints and $\Mb$ binary constraints, with $M=\Mm+\Mb$.

Using the aforementioned details, we discuss separately the benefits of \gpcr~in four different scenarios that arise from particular characterizations of the constraint and the objective. The explanations are supported with illustrations~(\cref{fig:applications}) on a simple one dimensional setting with one objective function and one constraint. In all cases, we also show the decision-making of the Bayesian optimizer that addresses the constrained minimization problem~\cref{eq:min_cost_cons}, and will be detailed in~\cref{sec:mesco}. As a quick reference for the reader, we briefly summarize here the four cases:
\begin{enumerate}[noitemsep,label=Case \arabic*),wide=0.0\parindent]
\small
	\item Single constraint equal to the objective; $M=\Mm=1$; self-constrained objective (\cref{ssec:selfcons}).
	\item Multiple binary constraints; $M=\Mb\geq 1$; constraints absorbed by the objective (\cref{ssec:multibin}).
	\item Multiple level-set constraints; $M=\Mm$; objective modeled as a standard GP (\cref{ssec:multimeas}).
	\item Self-constrained objective, and multiple level-set constraints; $\Mm \geq 1$ and $\Mb \geq 1$ (\cref{ssec:mixed}).
\end{enumerate}

\subsection{Case 1) Self-constrained objective}
\label{ssec:selfcons}
The general problem formulation~\cref{eq:min_cost_cons} finds its simplest case when the objective is also the constraint, i.e., $M=\Mm=1$, with $f(x)=g_1(x)$ and $c=c_1$. This case is equivalent to the one explained in~\cref{ssec:thres}: The threshold $c$ imposes a level-set constraint on $f$ (we say $f$ is \emph{self-constrained}), with stability region $\domsta = \left\lbrace x : g_1(x) \leq c_1 \right\rbrace = \left\lbrace x : f(x) \leq c \right\rbrace$. 
\cref{fig:applications:case1}~shows a possible situation corresponding to such a scenario. The objective is modeled with \gpcr, and the success/failure binary information is captured by it. The true constraint is also shown for clarity to emphasize that it is identical to the true objective, but it does not play any role. We see that \gpcr~pushes the probability mass above the current estimate of the instability threshold $\thresopt$, in areas with unstable evaluations. The first clear benefit of \gpcr~in this case is that no heuristic cost needs to be assigned to unstable controllers by the designer. The second, is that the Bayesian optimizer (bottom plot of \cref{fig:applications:case1})
sees those regions as less interesting because the cost is predicted to be high.

\subsection{Case 2) Multiple binary constraints}
\label{ssec:multibin}
\begin{figure}[h!]
\centering
\begin{subfigure}[b]{0.42\linewidth}
\includegraphics[width=\linewidth]{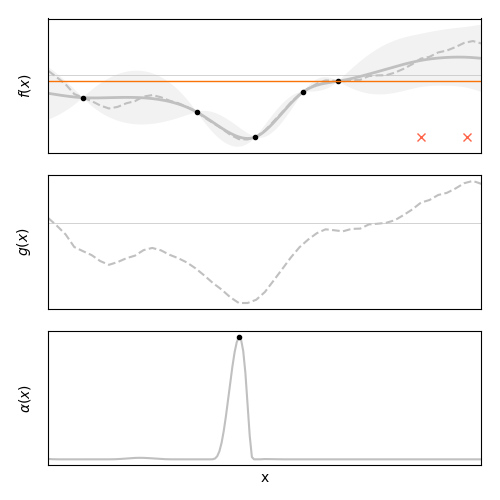}
\caption{Self-constrained objective}
\label{fig:applications:case1}
\end{subfigure}
\begin{subfigure}[b]{0.42\linewidth}
\includegraphics[width=\linewidth]{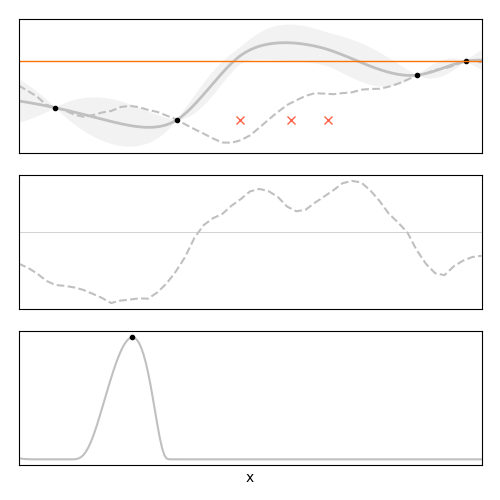}
\caption{Multiple binary constraints}
\label{fig:applications:case2}
\end{subfigure}

\begin{subfigure}[b]{0.42\linewidth}
\includegraphics[width=\linewidth]{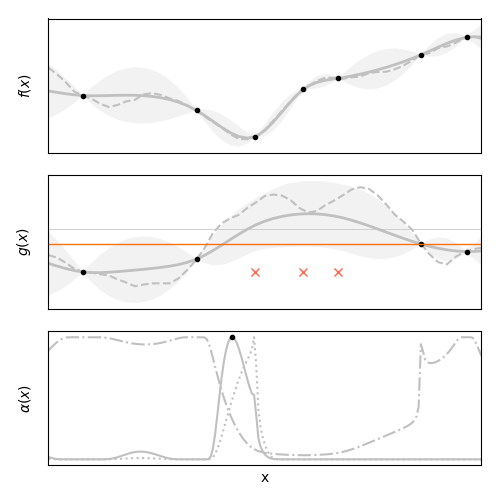}
\caption{Multiple level-set constraints}
\label{fig:applications:case3}
\end{subfigure}
\begin{subfigure}[b]{0.42\linewidth}
\includegraphics[width=\linewidth]{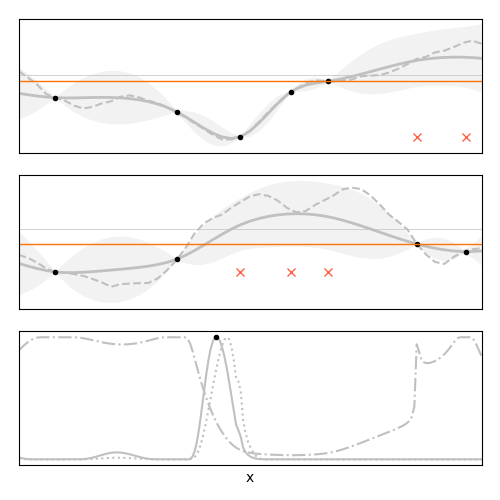}
\caption{Binary and level-set constraints}
\label{fig:applications:case4}
\end{subfigure}
\caption{One dimensional synthetic example that shows the use of the \gpcr~model in four different scenarios. For each scenario, the top plot shows the objective function $f$ and the middle plot shows the constraint $g$. When possible, the top and middle plots include the true function (dashed grey line) and the collected evaluations: stable (black dots) and unstable (red crosses). Also, when possible, the estimated instability threshold $\thresopt$ (orange horizontal line), and the true threshold $c$ (grey horizontal line) are also shown. The bottom plot shows the acquisition function (grey solid line) of the Bayesian optimizer. In plots (c) and (d), the probability of constraint satisfaction (dash-dotted line), and the unconstrained acquisition function (dotted line) are also shown as a reference. All three quantities are normalized to fit to the full range of the vertical axis.}
\label{fig:applications}
\end{figure}

In this case, we assume all constraints to be binary ($M=\Mb\geq 1$). Such constraint provides only discrete labels indicating success/failure. Thus, instead of modeling it separately (e.g., using a Gaussian process classifier), its discrete information is directly absorbed in the \gpcr~model used to model the objective.
In \cref{fig:applications:case2},~the region of stability is given by $\domsta = \left\lbrace x : g_1(x) \leq c_1 \right\rbrace$. When evaluating $f(x)$ in the middle region, where the constraint is violated, three unstable controllers are obtained. Therefore, the \gpcr~model pushes the probability mass above $\thresopt_1$. That way, those regions have a high probable cost.
The first clear benefit of \gpcr~in this case is that one model handles both, the information of the objective and the constraint.
The second clear benefit of \gpcr~is that it automatically makes the unstable region less interesting for the Bayesian optimizer, which selects the next evaluation in a stable area.
The estimated instability threshold $\thresopt_1$ is found above all collected stable points, but in this case it has no meaning beyond its role as a parameter within the \gpcr~model. It merely serves to push the cost probability above already collected stable evaluations.

\subsection{Case 3) Multiple level-set constraints}
\label{ssec:multimeas}
Herein, we assume all constraints to be level-set constraints ($M=\Mm$). 
 In \cref{fig:applications:case3}, the constraint $g_1$ is a level-set constraint, modeled with \gpcr, with the stable set determined by $\domstadef{g_1(x) < c_1}$, and $c_1$ being unknown a priori.
On the contrary, the objective function itself is not level-set constrained, and is modeled with a standard GP.
The Bayesian optimizer searches the constrained minimum $\fmincons$ in areas where the constraint is satisfied with high probability. The estimated instability threshold $\thresopt_1$ (orange horizontal line)
herein estimates the \emph{true} constraint threshold $c_1$ (grey horizontal line). The main benefit of \gpcr~in this case is that $c_1$ does not need to be known by the designer a priori. Instead, it can be estimated while collecting data.

\subsection{Case 4) Multiple binary and level-set constraints}
\label{ssec:mixed}
Last, we discuss the case in which $\Mm \geq 1$ and $\Mb \geq 1$.
Generally, we model each of the $\Mm$ level-set constraints independently with a \gpcr~model, and we consider the $\Mb$ binary constraints absorbed into the \gpcr~model for $f$. In total, this comprises $\Mm+1$ models. This is a more general case than the ones before, and mixes~\cref{ssec:multibin} and~\cref{ssec:multimeas} in one. \cref{fig:applications:case4} shows three main advantages of \gpcr~in this scenario: (i) The designer does not need to know a priori the threshold for the the level-set constraints, (ii) the binary constraints do not need to be modeled in addition, but they are rather absorbed by the model for $f$, and (iii) the designer does not need to define a penalty value for $f$.
It is worth remarking that this case is equivalent to having $\Mm \geq 1$ level-set constraints and a self-constrained objective. The reasoning is that, given a self-constrained objective, one can always construct a set of virtual binary constraints, which combined together, divide the space into the same stable/unstable areas implied by the instability threshold of the self-constrained objective.\\

\noindent We have illustrated the benefits of \gpcr~and its capability to handle multiple constraints in four different simple scenarios. However, it still remains a question how to solve \cref{eq:min_cost_cons} for each of the aforementioned four cases.
In the next section, we show how constrained Bayesian optimization can be leveraged to address \cref{eq:min_cost_cons}, and explain the nuisances for each of the four cases.

 \section{Constrained Bayesian optimization and \gpcr}
\label{sec:mesco}
Standard Bayesian optimization (BO) attempts to solve problem~\cref{eq:min_cost} by sequentially deciding which controller $\xnext$ shall be evaluated in the next experiment. Such decision is made based on a specific criterion, e.g., maximize information gain about the optimum~\citep{HennigS2012}, trade-off exploration vs.\ exploitation~\citep{srinivas2009gaussian}, or improve upon the best point observed so far~\citep{jones1998efficient}. For an extensive review on the recent BO methods, see~\citep{shahriari2016taking}. The aforementioned criteria are represented by an acquisition function $\acqui:\mathcal{X} \rightarrow \mathbb{R}$, typically computed through the Gaussian process model on $f$, updated with the function values revealed up to the current iteration. The maximizer of $\acqui$ is used to decide on the next experiment
$\xnext = \arg\max_{x\in\dom} \acqui(x)$. 

In presence of $M$ black-box constraints, \emph{Bayesian optimization with unknown constraints} (BOC) \citep{griffiths2017constrained,gelbart2015thesis,gardner2014bayesian,snoek2013bayesian,gelbart2014bayesian,schonlau1998global} has been proposed to address problem~\cref{eq:min_cost_cons}. Therein, the main idea is to target experiments where the constraints are satisfied with high probability.

In the following, we discuss the benefits of \gpcr~in the context of BOC and BO for the four cases described in~\cref{sec:applications}. In \cref{ssec:BOCbin} we discuss that, when all constraints are binary (i.e., \cref{ssec:multibin}), these can be absorbed into the \gpcr~model, and thus standard BO suffices to solve the constrained optimization problem.
Furthermore, in \cref{ssec:BOClevelset} we show that, when at least one of the constraints is a level-set constraint, we can incorporate it into the optimization problem, and address it using constrained Bayesian optimization. 

Although existing BOC methods have proven to be successful in practice \citep{gelbart2014bayesian}, the underlying decision-makers are heuristics developed based on intuition. Alternatively, entropy-based methods \citep{HennigS2012}, constructed over the principle of information gain about the optimum, have proven to outperform the aforementioned heuristics. However, entropy-based methods able to handle multiple constraints are scarce and computationally expensive, i.e., only \citep{lobato2016general}, to the best of our knowledge. Therefore, we propose to extend a recent entropy-based criterion \citep{wang2017mes}, which is computationally cheaper than \citep{lobato2016general}, to account for unknown constraints.

\subsection{Bayesian optimization with unknown level-set constraints}
\label{ssec:BOClevelset}
\cref{ssec:multimeas} and \cref{ssec:mixed} illustrate the case in which at least one of the $M$ constraints of~\cref{eq:min_cost_cons} is a level-set constraint, modeled with \gpcr. Since such model describes the constraint probabilistically, we cannot guarantee that the constraint will be satisfied/violated in locations $x$ where observations have not been collected. Furthermore, even at locations where the constraint is satisfied and a continuous-valued observation is acquired, we are unable to guarantee constraint satisfaction/violation, because the observation is contaminated with noise. As a result, it is not possible to ensure that the constraints are satisfied for any $x$. In light of these challenges, the minimizer of problem~\cref{eq:min_cost_cons} can by approximated by
\begin{equation}
\argmin_{x\in\dom}\; \E{f(x) | \sobs}, \text{ s.t. } \prod_{j=1}^M \Prob{\gn(x) \leq \gmaxn \mid \sobs_j} \geq \userconf,
\label{eq:min_cost_cons_boc}
\end{equation}
where $\userconf$ represents some user-defined level of confidence, with $\delta \in \left[0,1\right]$, and we have assumed that the constraints are modeled independently. This formulation has been proposed by~\citep{gelbart2014bayesian,gramacy2011opti}. The idea is to minimize the objective function in expectation while satisfying the constraints with high probability. The same authors also propose a powerful procedure to include the information provided by the probabilistic constraints into the exploration strategy given by the acquisition function $\acqui$. Although in their framework they consider expected improvement (EI)~\citep{mockus1978toward} as the acquisition function, they also state that their formulation is valid for any generic acquisition function $\acqui$. We use herein their formulation to propose a failures-aware acquisition function $\acquiFailOptiC$, which tends to explore in regions where the constraints are satisfied with high probability
\begin{equation}
\acquiFailOptiC(x) = 
\left\lbrace
\begin{array}{ll}
\acqui(x)\prod_{j=1}^M \Prob{\gn(x) \leq \hat{c}_j \mid \sobs_j}, & \text{if } \exists\;x, \prod_{j=1}^M\Prob{\gn(x) \leq \hat{c}_j \mid \sobs_j} \geq 1-\delta \\
\prod_{j=1}^M \Prob{\gn(x) \leq \hat{c}_j \mid \sobs_j}, & \text{otherwise}.
\end{array}
\right.
\label{eq:failoptic}
\end{equation}
The acquisition \cref{eq:failoptic} has two modes of operation: (a) If there exists at least one $x$ for which all constraints are satisfied with high probability, we downsize $\alpha(x)$ with the probability of constraint satisfaction (i.e., points that are more likely to satisfy the constraints will get a higher weight); (b) if the constraints are violated for all $x$ with high probability, whatever point is suggested by $\acqui(x)$ will violate the constraint as well. In this case, it is better to exclude $\acqui(x)$ and select the point that has the highest probability of satisfying the constraint~\citep{griffiths2017constrained}. The latter is merely an exploratory strategy that will seek for some region where the constraint is satisfied. Once such a region is found, the strategy switches to (a).

An important difference of~\cref{eq:failoptic} with respect to~\citep{gelbart2014bayesian} is that we assume the true constraint thresholds $c_j$ to be unknown. Instead, we consider the estimated thresholds $\hat{c}_j$, available through the \gpcr~model (see \cref{ssec:threshold}). In fact, when modeling $g_j$ with \gpcr, the probability $\Prob{\gn(x) \leq \hat{c}_j \mid \sobs_j}$ is a natural consequence of the model and can be computed following~\cref{eq:predictive_constraint}.

\cref{fig:applications:case3,fig:applications:case4} show the operation mode (a) of the acquisition function $\acquiFailOptiC$ in light of the constraint $g$. In both scenarios, the probability of constraint satisfaction $\Prob{g(x) < \hat{c}_g \mid \sobs_g}$, which multiplies the unconstrained acquisition function $\alpha$, are also shown for clarity. In both cases it is clear that downsizing $\alpha$ by the probability of constraint satisfaction changes the decision making to areas where it is less likely to violate the constraint.

\subsection{Bayesian optimization with unknown binary constraints}
\label{ssec:BOCbin}
Herein, we consider the case in which all the constraints $g_j$ from~\cref{eq:min_cost_cons} are binary. As detailed in~\cref{ssec:multibin}, we propose to absorb them by the \gpcr~model of $f$. This makes the model push up the probability of the cost at regions predicted to be unstable. Then, those regions will automatically look less appealing to the acquisition function of standard BO algorithms. Therefore, it becomes sufficient to deploy standard BO instead of BOC, although the problem~\cref{eq:min_cost_cons} is a \emph{constrained} optimization problem. This is the main benefit of using the \gpcr~model in presence of only binary constraints.

Interestingly enough, one could anyways use the BOC strategy~\cref{eq:failoptic} in this scenario by considering $\Prob{f(x) < \hat{c} \mid \sobs_g}$ as the only probabilistic constraint. However, in our experiments such constraint seemed to be redundant, since regions where $\Prob{f(x) < \hat{c} \mid \sobs_g}$ is small are anyway rarely explored.

The acquisition function $\alpha$ in~\cref{eq:failoptic} is a placeholder that can be replaced by any improvement criterion~\citep{gelbart2014bayesian}, in any problem with level-set constraints. In the following, we propose a Bayesian optimization criterion that accounts for unknown level-set constraints.

\subsection{Min-Value Entropy Search with Unknown Level-Set Constraints (\mesco)}
Max-Value Entropy Search~\citep{wang2017mes} is a Bayesian optimizer that addresses the problem~\cref{eq:min_cost} from an information-theoretic perspective. Whereas~\citep{wang2017mes} consider a maximization problem in their formulation, we characterize it from the point of view of a minimization problem (\mes), and extend it to account for unknown level-set constraints (\mesco). The resulting algorithm is called \emph{min-Value Entropy Search with Unknown Level-Set Constraints} (\mesco). In the following, we first introduce \mes~and then explain the pertinent modifications that lead to \mesco.

\subsubsection{\mes}
The acquisition function proposed by \mes~is defined as the gain in mutual information between the next point to query $\query$ and the minimum $\fmin$
 \begin{equation}
\begin{aligned}
\acquimes(x) & = \MI{\query}{\fmin | \sobs} \\
& = H(\prob{f|\sobs,x}) - \mathbb{E}_{\fmin \sim p(\fmin|\sobs)} \left[ {H(\prob{f|\sobs,x,\fmin})} \right] \\
& \simeq \dfrac{1}{S}\sum_{i=1}^S\left[  -\dfrac{z_i(\inx)\phi \left( z_i(\inx) \right) }{2\cdf{-z_i(\inx)}} -\log \cdf{-z_i(\inx)} \right],
\label{eq:mES}
\end{aligned}
\end{equation}
where $\Phi$ is the cumulative density function, $\phi$ is the density function of a normal distribution, and $z_i = (\fmin^i-\mu(x))/\sigma(x)$. 
The distribution $\prob{f|\sobs,x,\fmin}$ is a truncated Gaussian with support $(\fmin,+\infty)$. 
The acquisition function reveals areas where there is significant variance falling below $\fmin$. Acquiring $S$ samples $\fmin^i \sim p(\fmin|\sobs)$ is challenging because the distribution $p(\fmin|\sobs)$ is unknown a priori. To this end, \citep{wang2017mes} propose two different approaches to obtain samples from $p(\fmin|\sobs)$, which involve numerical approximations that will be briefly explained in the following, while explaining \mesco, for convenience.\\

\noindent Extending \mes~to account for $M$ unknown constraints requires implementing some modifications to the original algorithm, which we detail next.

\subsubsection{\mesco}
\label{sssec:mesco}
When addressing the constrained optimization problem \cref{eq:min_cost_cons_boc}, \mes~cannot be used directly because: (i) it gathers samples from the unconstrained minimum $\fmin^i$, while we are interested in the samples of the constrained minimum $\fmincons^i \sim p(\fmincons|\sobs)$, and (ii) using directly samples $\fmincons^i$ on \cref{eq:mES} makes the acquisition function have extremely large peaks when the sample value $\fmincons^i$ is near the values of the observations. 
For these reasons, we modify the original $\acquimes$ strategy by applying the following steps:
\begin{enumerate}
\item Since the true constraint thresholds $\thres_j$ are unknown, we update the current approximates $\hat{c}_j$, given by the \gpcr~model of each constraint $g_j$, as indicated in \cref{sssec:MAP}.
\item In \citep{wang2017mes}, it is proposed to either approximate $p(\fmin|\sobs)$ using a Gumbel distribution, which involves discretizing the input domain $\dom$, or 
approximate 
random function realizations of the GP posterior and minimize them using local optimization. The approximation in the latter proposition comes from applying Bochner's theorem, which provides a callable object, but as a counterpart, it creates undesired harmonics in the approximated realization that can mislead the search. Additionally, it requires computing the spectral density of the kernel, which might not always be available (see \citep{hernandez2014predictive,wang2017mes} for details).

While \citep{wang2017mes} sample from $p(\fmin|\sobs)$, we need to sample from the distribution over the constrained minimum $p(\fmincons|\sobs)$. For this, we follow a method that does not require numerical approximations. Herein, we describe how to obtain one sample $\fmincons^i$ by running local optimization with random restarts on the problem \cref{eq:min_cost_cons} using the approximated thresholds $\hat{c}_j$, and virtual evaluations. Every time the local optimizer requests an observation of $f$ or $g_j$ at a location $x$, we
\begin{enumerate}[label=\alph*)]
\item sample such observation from the predictive distribution of the corresponding GP, i.e., $f_r \sim p(f | \sobs,x)$, which moments are given in \cref{eq:pred_q_moments},
\item contaminate it with Gaussian noise, $y_r  = f_r + \varepsilon,\; \varepsilon \sim \N{\varepsilon}{0,\sigma^2}$, and
\item include the sampled point in a virtual dataset $\mathcal{I}_{r+1} \gets \mathcal{I}_r \cup \{ x , y_r \}$, which is initialized with the current set of $N$ evaluations, i.e., $\mathcal{I}_0 \gets \sobs$ and expanded as virtual evaluations are collected.
\end{enumerate}
This operation is repeated for every new point the local optimizer request in each one of the functions $f$, $g_j$. Collecting virtual evaluations in this way is equivalent to collecting them from a pre-sampled realization of the GP, with the advantage that it does not need numerical approximations. 
\noindent The caveat of this approach is that the prior kernel matrix $\tilde{\Sigma}$ from \cref{eq:my_post} needs to be augmented and inverted every time a new virtual evaluation is added, which has a maximum cost of $O( \sum_{r=1}^R (N+r)^3 )$, where $R$ is the maximum number of evaluations we allow per random restart. To alleviate this problem, we expand the inverse of the prior covariance using the Woodbury identity \citep[A.3]{Rasmussen2006Gaussian}, which has maximum cost of $O(RN^2)$.

\noindent 
Once the local optimization has finished, we return the constrained optimum $\fmincons^i$ as a sample. For each random restart, we reset $\tilde{\Sigma} \leftarrow \tilde{\Sigma}_\text{ext}$ and repeat this operation until $S$ samples are collected.
\item We modify \cref{eq:mES} to include the samples from $p(\fmincons|\sobs)$ by replacing $z_i$ with $\tilde{z}_i = (\fmincons^i-\mu(x))/\sigma(x)$
\item We drive the acquisition function toward regions that satisfy the constraints with high probability following the strategy~\cref{eq:failoptic}, where $\alpha$ is to be replaced by $\acquimesco$. As a side effect, the peaks originated by samples $\fmincons^i$ being near existing evaluations are effectively downsized by the probability of success, which is low in unsafe areas, where the peaks typically occur. This, in practice they only do not affect the decision-making.
\end{enumerate}

\noindent The resulting acquisition function is smooth and can be maximized using analytical gradients:
\begin{equation}
\acquimesco(x) = 
\left\lbrace
\begin{array}{ll}
\acquimes(x)\prod_{j=1}^M \Prob{\gn(x) \leq \hat{c}_j \mid \sobs_j}, & \text{if } \exists x,\; \prod_{j=1}^M\Prob{\gn(x) \leq \hat{c}_j \mid \sobs_j} \geq 1-\delta \\
\prod_{j=1}^M \Prob{\gn(x) \leq \hat{c}_j \mid \sobs_j}, & \text{otherwise}.
\end{array}
\right.
\label{eq:mesco}
\end{equation}
A pseudocode for \mesco\footnote{Our python implementation of both, \gpcr~and \mesco~will be made available upon publication.}~is presented in \cref{alg:mESCO}, and additional implementation details are given in \cref{app:mescoalg}. An intuitive idea of the decision-making of \mesco~is visually illustrated in the one-dimensional synthetic examples (\cref{fig:applications:case3,fig:applications:case4}). A more thorough analysis in different benchmarks and dimensionality is provided in \cref{sec:exper}.

It is worth remarking that the proposed algorithm is not tied to the use of \gpcr. For example, if the constraints are not level-set constraints (i.e., unknown thresholds), but standard constraints (i.e., known thresholds), then all the \gpcr~models can be replaced by standard GPs, and \cref{alg:mESCO} can equally be used by simply skipping the update steps from lines 3 and 4, and using the true constraint thresholds instead.

 \section{Related work and contributions}
\label{sec:relWork}

There has been recent interest in Bayesian optimization (BO) \citep{shahriari2016taking} for learning robot controllers~\citep{vonrohr2018gait,rai2017bayesian,marco16tuning,calandra2016bayesian,Berkenkamp2016SafeOpt}. While in these works, unstable evaluations have been included in the data set during learning in form of a high penalty, in this paper this issue is fundamentally addressed using \gpcr. In the following, we show close connections between the core ideas of this work with other methodologies similar or derivate from BO.

A similar methodology to BO is active learning, where the next controller is selected using an acquisition criterion. For example, in~\citep{schreiter2015safe}
the optimization problem~\cref{eq:min_cost_cons} is approached using safe exploration.
More specifically, a negative label is assigned to failure cases, and a positive label is assigned to success cases. A Gaussian process classifier (GPC) is used to learn a probabilistic binary constraint that allows for making probabilistic statements about whether a region is likely to be stable or unstable. While this method appears to be simple, it involves carrying two models: One Gaussian process regression model (GPR) for the objective, and one GPC to model the constraint.
The authors show that the input space is safely explored, while the unknown constraint $g$ is learned alongside.
Observations of such constraint are modeled with a heterogeneous likelihood that perceives (i) continuous valued observations, near the stability boundary, and (ii) discrete labels (unstable/stable), far from it. While this approach shares similarities with the one presented herein, we propose a Bayesian model that explicitly bypasses the need of an extra GPC model by including the discrete labels directly in the GPR, as detailed in in~\cref{ssec:GP}.
Such model is mathematically different from that of~\citep{schreiter2015safe} and carries less hyperparameters to estimate or choose, while it retains the same flexibility.
Finally, the authors~\citep{schreiter2015safe} assume that continuous valued observations of $g$ are only possible near the transition boundary, while we do not need that assumption.

Another closely related work~\citep{gotovos2013active} in the context of active learning proposes a method that solves a classification problem for function values to lie above or below a specific threshold. Such problem is posed as a level set estimation problem, where the threshold is either explicitly given by an expert, or implicitly defined within a confidence interval. In the latter case, the authors give convergence guarantees for such interval to decrease below a desired accuracy after a number of evaluations. 
While our problem formulation shares similarities with this one, there are fundamental differences: (i)
Their 
target is to classify the input space in stable vs. unstable areas, while ours is to find the optimum avoiding unstable regions, revealed during exploration; (ii)
in their formulation, the threshold does not impose a critical constraint and thus, observations of the objective can be obtained above it. In our formulation, constraint violation is critical in the sense that no observations of the objective are obtained above the threshold, but just a label indicating that the evaluation was unstable.

In the context of RL for robotics, BO has also been combined with policy search methods to learn feedback policies \citep{englert2018learning,driess2017constrained}.
Therein, the unknown constraint is modeled as a GPC, which is then used to compute a safe boundary region in the input domain. They include the information of the boundary directly into the Bayesian optimization criterion, which then trades off exploration, near the current boundary estimate, with exploitation, in a region inside the boundary estimate. While such method is relevant in the context of safe learning, we observe two main aspects: (i) it needs a stable policy as a starting point, and (ii) it expands the safe region around this inital controller, leaving unexplored potential better optima that could exist outside of it. Instead, our method does not assume a single safe region that gets expanded, but can identify multiple safe regions for which the probability of constraint satisfaction is high. Additionally, although they propose safe exploration, their method is not exemt of failures when identifying the safety boundary on a real robot. When failures occur, we see them as a benefit, since they are informative about regions that shall not be visited again. The interesting case in which the optimal solution lies close to the boundary can be found by our method, as well.

Bayesian optimization with unknown constraints (BOC)~\citep{gardner2014bayesian,snoek2013bayesian,gelbart2014bayesian,gramacy2011opti,picheny2014stepwise,schonlau1998global,schreiter2015safe,griffiths2017constrained,lobato2016general} has shown to be effective in well-known optimization benchmarks~\citep{gelbart2015thesis,gelbart2014bayesian} and in applications such as learning the hyperparameters of a neural network or Markov chain Monte Carlo methods~\citep{lobato2016general}. The target of the constrained optimization problem \cref{eq:min_cost_cons} is to optimize an unknown objective while satisfying multiple unknown constraints with high probability. 
When it comes to determine constraint satisfaction, there are two main differences between the aforementioned BOC techniques and our approach:
First, while in BOC it is typically assumed that the constraint threshold $\gmaxn$ is known a priori
\footnote{While in the aforementioned papers, the constrained optimization problem is generally formulated as $\min_{x\in \mathcal{X}}f(x),\; \text{s.t. } \tilde{g}_1(x) \geq 0, \dots, \tilde{g}_M(x) \geq 0$, we can show that our formulation~\cref{eq:min_cost_cons} is equivalent by defining $\tilde{g}_j(x) = \gmaxn - \gn(x)$. It is important to notice that the constraint $\tilde{g}_j(x) \geq 0$, as presented in BOC, implicitly assumes that the threshold (zero) is known, while we do not assume so.}
, we do not assume so.
 In our work, we explicitly emphasize a generic unknown threshold $\gmaxn$, which does not need to be defined a priori, but can be learned from data, either estimated as a model parameter, or by treating it from a Bayesian perspective. This aspect is a core contribution of this work, novel in the context of BOC, and not explored by the aforementioned papers.
Second, a prevailing assumption in BOC is that 
noisy observations of the constraints are always available, even after constraint violation, while in this work, we assume that no continuous-valued observation is obtained upon constraint violation, as stated in~\cref{sec:applications}.
Alternatively, we assign to those measurements a negative label, which is captured by the proposed Bayesian model, and allows to make inference and improve the knowledge about the unknown threshold.

\noindent To summarize, in 
this work we propose a general Bayesian model that can be used to model objective and constraints in BO when unstable evaluations yield no continuous-valued observations. In addition, we extend an existing BO method to account for unknown constraints. In the context of the related work, a detailed list of the contributions is summarized below.
\begin{enumerate}
	
	\item \textbf{One model for both, classification and regression:} 
			The proposed \gpcr~model bypasses the need of training an additional GP classifier, as done in~\citep{schreiter2015safe,snoek2013bayesian,englert2018learning,driess2017constrained}, resulting in an equally flexible model, but with a lower number of parameters.

	\item \textbf{No failure penalties, and no constraint thresholds:}
		The proposed Bayesian model circumvents the need of defining an arbitrary penalty a priori for failures, as commonly done in RL and BO. Instead, whenever the roll-out becomes unstable, the model pushes the predictive probability mass above the unknown stability threshold $\thres$. In that way, failure points and their vicinity have lower chances to be re-visited again during the optimization search. The unknown stability threshold $\thres$ is a key parameter of the model, and is re-learned from data at each iteration, instead of being heuristically chosen by a designer. 

	When additional indicators of failure are accessible, they can be included in the problem as \emph{additional} constraints $\gn$. Such constraints can then be modeled using the same aforementioned model, for which the constraint threshold $\threscons$ is unknown, but can also be learned from data.
	
	To the best of our knowledge, \gpcr~is the first Bayesian model able to estimate the threshold from a hybrid data set that combines discrete labels and observations of the objective, in the context of GP regression and Bayesian optimization.  

					\item \textbf{Min-Value Entropy Search with unknown level-set constraints (\mesco):} Besides the \gpcr~model, we propose a novel extension of a recent entropy-based Bayesian optimizer: Max-Value Entropy Search~\citep{wang2017mes} to account for unknown constraints: \mesco. Such method is used to validate \gpcr, first in several synthetic benchmarks, and second, using a humanoid robot balancing an inverted pole as experimental platform.

\end{enumerate}
 \section{Experimental results}
\label{sec:exper}
In this section, we validate \gpcr~in the context of black-box constrained optimization. We asses the capability of the model to learn the objective and the constraints when their corresponding thresholds are unknown. At the same time, we test the performance of \mesco~in finding the constrained global minimum. To this end, we present three different settings. First, in \cref{sec:exper_illus}, we illustrate the benefits and caveats of \gpcr~in finding the global minimum in two simulated optimization problems. Second, in \cref{sec:statistical}, we report on the consistency of \gpcr~and our implementations of \mes~and \mesco~by benchmarking the same simulated scenarios. Last, in \cref{sec:robexp}, we show the benefit of \gpcr~and \mesco~in finding the controller parameters of a real robot balancing an inverted pole.

\subsection{Illustrative examples}
\label{sec:exper_illus}
In this section, \gpcr~is demonstrated in two different simulated scenarios in 2D. The first simulation illustrates the usability of \gpcr~when the objective is self-constrained (cf. \cref{ssec:selfcons}), and the second simulation shows the case of one unconstrained objective and one level-set constraint (cf. \cref{ssec:multimeas}).

\subsubsection{Experiment design choices}
For both simulations, the domain of the objective functions and the constraints is scaled to the unit square $\dom = \left[ 0,1 \right]^2$. 
We consider an isometric Mat\'{e}rn 3/2 kernel, and all lengthscales and variances fixed a priori. In both cases, we assume a Gaussian hyperprior $p(\thres)=\N{\thres}{0,\sigma_\thres^2}$ over the instability threshold in Simulation I and over the constraint threshold in Simulation II. Such thresholds are re-estimated after each iteration using MAP, as explained in~\cref{sssec:MAP}. The user confidence level is fixed with $\delta=0.05$ for all the experiments. The initial point $x_0$ was randomly sampled from a uniform distribution for all cases. For all evaluations, we consider small evaluation noise, i.e., $\sigma = 0.01$.

\subsubsection{Simulation I}
\begin{figure}[H]
\begin{subfigure}{0.24\linewidth}
\includegraphics[width=\linewidth]{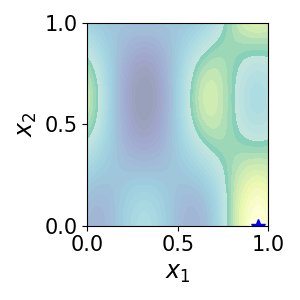}
\caption{$f(x_1,x_2)$}
\label{sfig:illus1_gardner}
\end{subfigure}
\begin{subfigure}{0.24\linewidth}
\includegraphics[width=\linewidth]{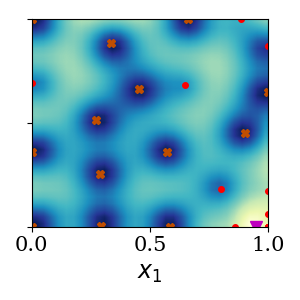}
\caption{\gpcr~mean}
\label{sfig:illus1_mean}
\end{subfigure}
\begin{subfigure}{0.24\linewidth}
\includegraphics[width=\linewidth]{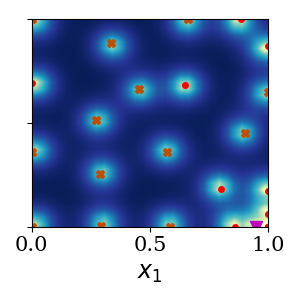}
\caption{\gpcr~std. deviation}
\label{sfig:illus1_std_dev}
\end{subfigure}
\begin{subfigure}{0.24\linewidth}
\includegraphics[width=\linewidth]{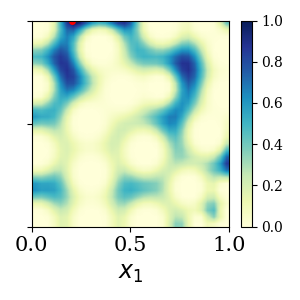}
\caption{\mesco}
\label{sfig:illus1_mesco}
\end{subfigure}
\caption{
Global search on the objective function \cref{eq:gardner_fun} using \mes. (a) shows the unstable region $\domuns$ (shadowed area), and the true feasible global minimum (blue star). (b) and (c) show the predictive mean and standard deviation of \gpcr, respectively, stable evaluations (red dots) unstable evaluations (orange crosses), and the best guess estimation of the global minimum $\xbg$ (magenta triangle). (d) shows the \mes~acquisition function, and the next suggestion $\xnext$ (red dot). Subsequent figures in the paper follow the same legend.}
\label{fig:illus_gardner}
\end{figure}

We consider the 2D objective function
\begin{equation}
f(x_1,x_2) = \cos\left( 10x_1\right)\cos\left( 5x_2\right) + \sin\left( 10x_1\right) + 2,
\label{eq:gardner_fun}
\end{equation}
proposed in~\citep{gardner2014bayesian} as a benchmark to test constrained Bayesian optimization. While they consider $f$, together with an external constraint $g$, we decide instead to consider that $f$ is self-constrained (cf. \cref{ssec:selfcons}), with instability threshold $\thres=1.5$. This induces a region of unstable controllers $\domuns= \left\lbrace x : f(x_1,x_2) > \thres \right\rbrace $, depicted in \cref{sfig:illus1_gardner} as a shadowed area. We set a wide hyperprior on the threshold, i.e., $\sigma_\thres=5$, and run \mes~for 30 iterations.
\cref{sfig:illus1_mean,sfig:illus1_std_dev} show that \mes~finds the optimal value after 17 stable and 13 unstable evaluations. The last 12 evaluations are stable and targeted in the area around the global minimum. At the final iteration, the estimated best guess is $\xbg=\left(0.9468,0\right)$, $f(\xbg)=0.001177$, while the true global minimum is $\xtrue=(0.942,0)$, $f(\xtrue)=0$. The instability threshold is estimated to be $\thresopt=1.35$.
While in~\citep{gardner2014bayesian}, an algorithm for constrained Bayesian optimization is proposed, it is not possible to fairly compare our framework with theirs because we define $g$ differently (i.e., we consider constraint and objective identical). However, we can highlight two clear benefits of using \gpcr. First, 
the \gpcr~model on the objective absorbs the failures, while in~\citep{gardner2014bayesian}, any existing black-box constraint needs to be modeled with an additional standard GP. Furthermore, we can search for the global minimum using an unconstrained Bayesian optimization method, such as \mes, without the need of using a constrained Bayesian optimizer (like the one proposed in~\citep{gardner2014bayesian} or \mesco).

\subsubsection{Simulation II}
\begin{figure}[H]
\begin{subfigure}[b]{0.24\linewidth}
\includegraphics[width=\linewidth]{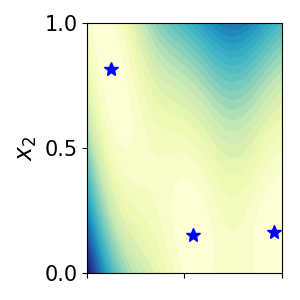}
\caption{Branin}
\label{sfig:illus2_branin}
\end{subfigure}
\begin{subfigure}[b]{0.24\linewidth}
\includegraphics[width=\linewidth]{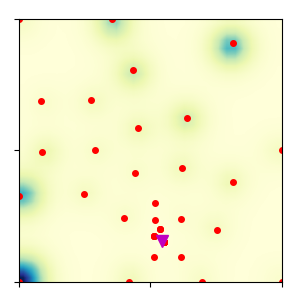}
\caption{\gpcr~mean}
\label{sfig:illus2_mean}
\end{subfigure}
\begin{subfigure}[b]{0.24\linewidth}
\includegraphics[width=\linewidth]{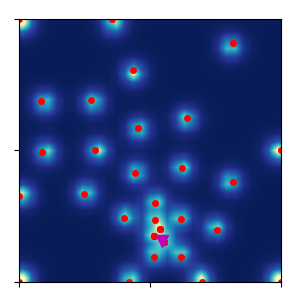}
\caption{\gpcr~std. deviation}
\label{sfig:illus2_std_dev}
\end{subfigure}
\begin{subfigure}[b]{0.24\linewidth}
\includegraphics[width=\linewidth]{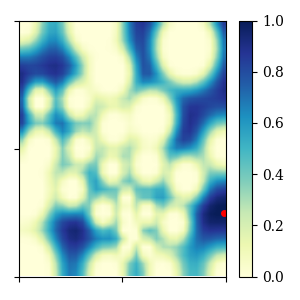}
\caption{$\text{acqui}(x)$}
\label{sfig:illus2_mesco}
\end{subfigure}\\
\begin{subfigure}[b]{0.24\linewidth}
\includegraphics[width=\linewidth]{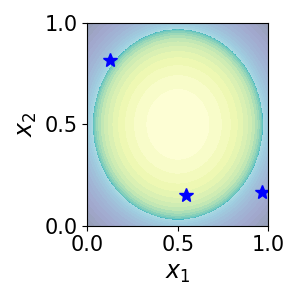}
\caption{Circle}
\label{sfig:illus2_circle}
\end{subfigure}
\begin{subfigure}[b]{0.24\linewidth}
\includegraphics[width=\linewidth]{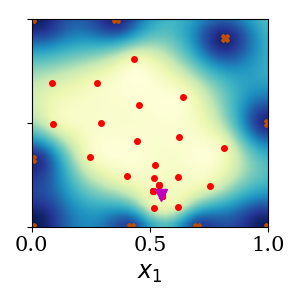}
\caption{\gpcr~mean}
\label{sfig:illus2_circle_mean}
\end{subfigure}
\begin{subfigure}[b]{0.24\linewidth}
\includegraphics[width=\linewidth]{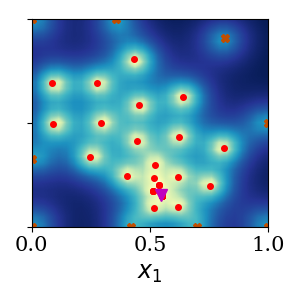}
\caption{\gpcr~std. deviation}
\label{sfig:illus2_circle_std_dev}
\end{subfigure}
\begin{subfigure}[b]{0.24\linewidth}
\includegraphics[width=\linewidth]{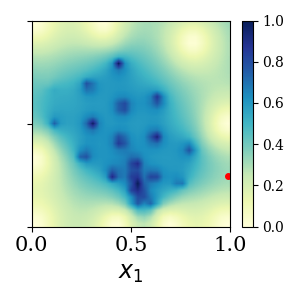}
\caption{$\Pr{g(x)\le \thresopt}$}
\label{sfig:illus2_circle_mesco}
\end{subfigure}
\caption{
Global search on the objective function \cref{eq:branin_fun} using \mesco. (a) and (e) show the objective function $f$ and the constraint $g$ from~\cref{eq:branin_fun}, respectively. In (e), the shadowed area represents constraint violation. (b),(c) show the predictive mean and std. deviation for $f$, and (f),(g) for $g$. (d) shows the acquisition function and (h) shows the probability of constraint satisfaction using the current estimate $\thresopt_1$ of the true constraint threshold $\thres_1$.
  }
\label{fig:illus_branin}
\end{figure}

We consider the well-known 2D Branin-Hoo benchmark, constrained to a centered circle
\begin{align}
f(x_1,x_2) & = (15x_2 - \dfrac{5.1}{4\pi^2}(15x_1-5)^2 + \dfrac{5}{\pi}(15x_1-5) - 6 )^2 + 9.6\cos(15x_1-5) + 10)^2 \nonumber \\
g(x_1,x_2) & = -\sqrt{\sfrac{2}{9}-(x_1-0.5)^2 - (x_2-0.5)^2} \leq \thres_1,
\label{eq:branin_fun}
\end{align}
with $\thres_1 = 0$, which has also been used in~\citep{gelbart2014bayesian}. The constraint violation region is depicted as a shadowed area in \cref{sfig:illus2_circle}, and the objective is shown in \cref{sfig:illus2_branin}. The objective has three minima, marked in \cref{sfig:illus2_branin,sfig:illus2_circle}, but the constraint hides two of them, leaving only one as the solution to the constrained optimization problem. The squared root in the constraint function $g$ is deliberately introduced to make it undefined in the outer part of the circle, i.e., if violated, no continuous-valued observation can be obtained, but only a discrete label.
This fact characterizes $g$ as a level-set constraint (cf. \cref{ssec:multimeas}). We model $g$ using \gpcr, while $f$ is modeled with a standard GP.
We have chosen a wide hyperprior on $\thres_1$, with $\sigma_{c_1}=2.0$. 
\cref{fig:illus_branin} shows our proposed constrained Bayesian optimization algorithm, \mesco,~after 50 iterations, from which 9 were unstable and 41 were stable. At the final iteration, the estimated best guess is $\xbg=\left(0.5448,0.1527\right)$, $f(\xbg)=0.4039$, while the true constrained minimum is $\xtrue=(0.5428,0.1517)$, $f(\xtrue)=0.40$. The instability threshold is estimated to be $\thresopt_1=-0.09842$. These results are comparably better than~\citep{gelbart2014bayesian}, where their estimated global minimum is reported at $\left( 0.5340,0.1573 \right)$, with value $0.48$, and known threshold. Furthermore, the advantage of using \gpcr~on the constraint as compared to a standard GP classifier~\citep{gelbart2014bayesian} is that \gpcr~can afford the constraint threshold $\thres_1$ being unknown a priori, and it can be estimated over iterations.

\subsection{Statistical comparison}
\label{sec:statistical}

After having shown the general applicability of the proposed framework, we now report on the consistency of \gpcr~and our implementations of both, \mes~and \mesco. To this end, we run the experiments from the previous section multiple times and measure
the performance of the global optimization methods using the inference regret $R_i$ at each iteration $i$, defined as
$R_i = y(x_{\text{bg},i}) - \min_{x\in \domsta} f(x)$ \citep{wang2017mes}, where $y(x_{\text{bg},i})$ is a function evaluation at the best guess at iteration $i$. 

\begin{figure}[H]
\centering
\begin{subfigure}{0.40\linewidth}
\includegraphics[width=\linewidth]{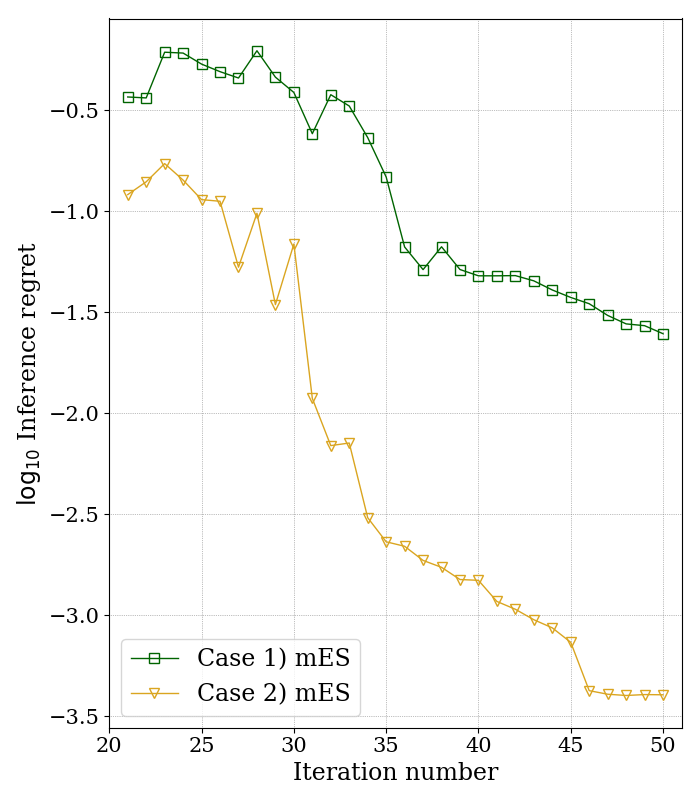}
\caption{mES}
\label{sfig:statistical12}
\end{subfigure}
\begin{subfigure}{0.40\linewidth}
\includegraphics[width=\linewidth]{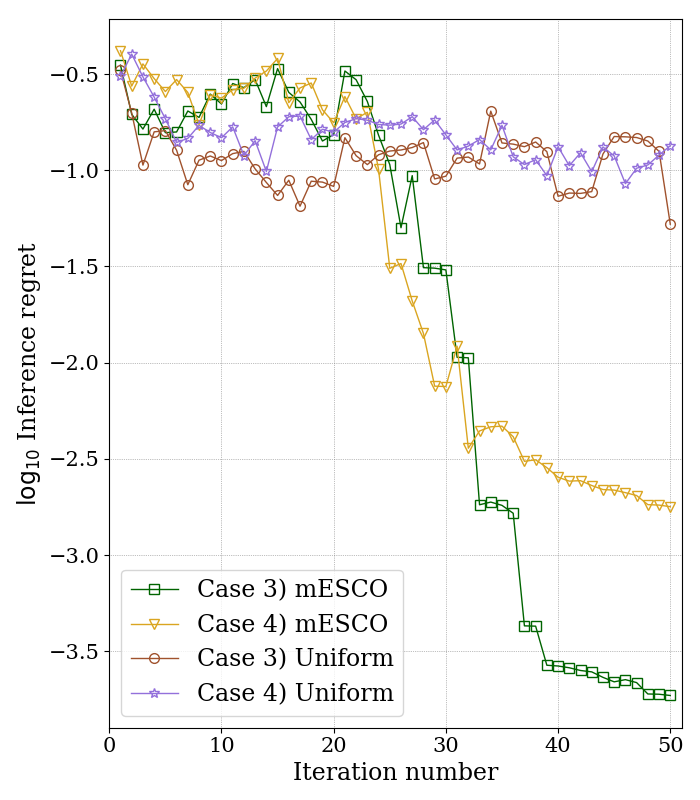}
\caption{mESCO}
\label{sfig:statistical34}
\end{subfigure}\\
\begin{subfigure}{0.8\linewidth}
\includegraphics[width=\linewidth]{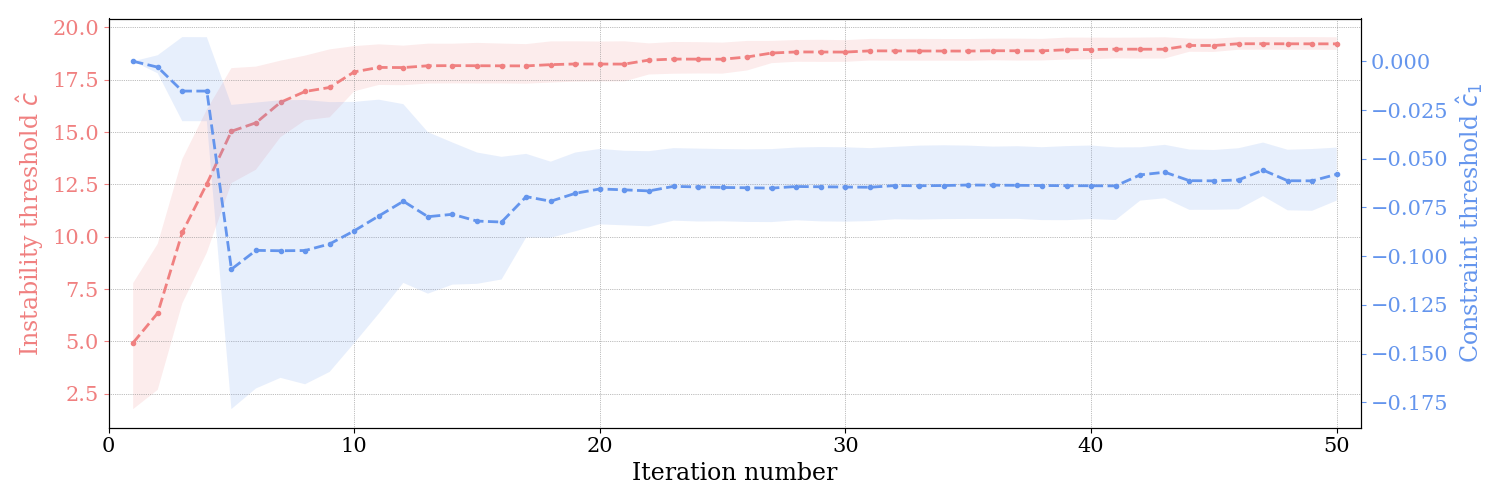}
\caption{Thresholds}
\label{sfig:statistical_thres}
\end{subfigure}
\caption{Statistical comparison}
\label{fig:statistical}
\end{figure}

We benchmark the experiments described in \cref{sec:exper_illus} over 20 runs for each case and report the averaged $R_i$ in \cref{sfig:statistical12} and \cref{sfig:statistical34}. We can see the results from benchmarking Simulation I in the green curve in \cref{sfig:statistical12}, and the results from benchmarking Simulation II in the green curve in \cref{sfig:statistical34}, which correspond to cases 1) and 3) described in \cref{ssec:selfcons} and \cref{ssec:multimeas}, respectively. The remaining cases 2) and 4) are also addressed herein, by reusing the benchmark on Simulation II, but with a few variants, depending on the case.

\noindent To address case 2) (\cref{ssec:multibin}), we assume that $g$ is a binary constraint, and that its binary information is absorbed by \gpcr, used to model the objective $f$. To address case 4) (\cref{ssec:mixed}), we assume that $f$ is self-constrained, with instability threshold $c = 20$, and $g$ is a level set constraint.
\cref{sfig:statistical12} shows the performance of \mes~in case 2), and \cref{sfig:statistical34} shows the performance of \mesco~in case 4). To asses the performance of \mesco, we compare against random search using uniform sampling in \cref{sfig:statistical34}. As shown, \mesco~achieves the lowest regret in case 3). We also see that \mesco~achieves better minima than the one reported in~\citep{gelbart2014bayesian}, on average, i.e., $f(\xbg)=0.4717 \pm 0.1392$. 
For all cases, the solution is approximately found with a low error. 

\noindent In \cref{sfig:statistical_thres}, we depict the evolution of $\thresopt$ and $\thresopt_1$ over iterations for case 4). 
While initially the estimation shows a broad transient, it eventually exhibits a convergent behavior, obtaining on average $\thresopt=19.2459\pm 0.6714$ and $\thresopt_1=-0.0618\pm 0.0339$ at the last iteration. This shows that only a rough approximation to the true threshold is needed for \gpcr~to work in practice, when using it for BO.

\subsection{Robot experiments}
\label{sec:robexp}
\begin{wrapfigure}{R}{0.6\linewidth}
\centering
\includegraphics[width=0.6\linewidth]{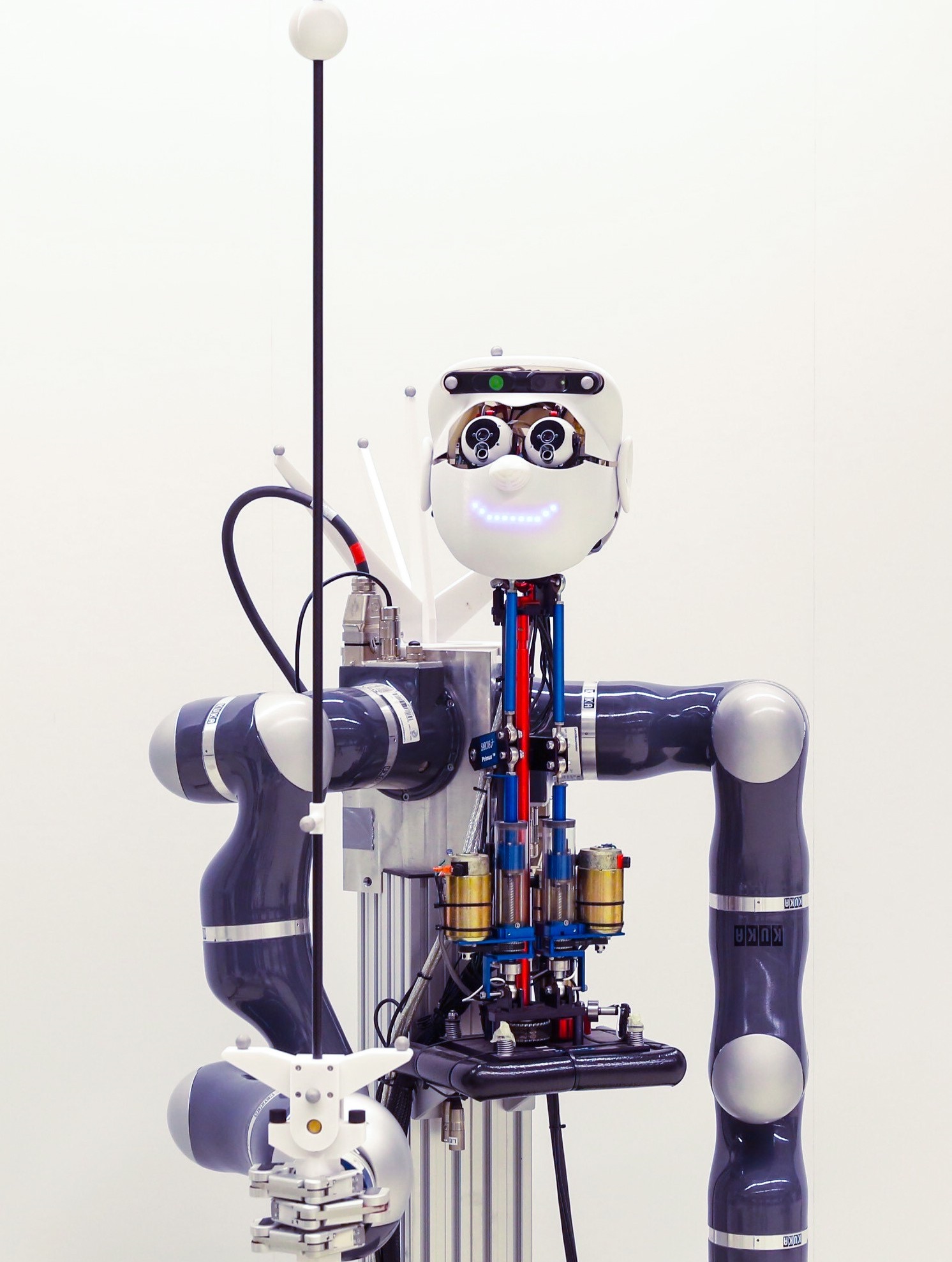}
\caption{Robot Apollo balancing an inverted pole}
\end{wrapfigure}
The simulation study revealed the possible benefits of \gpcr~and the effectiveness of \mesco~in synthetic scenarios. In this section, we will evaluate its performance on a real robotic platform: The humanoid robot Apollo balancing an inverted pole. 
This is a much more challenging scenario than the aforementioned simulations because (i) the latent objective and constraint functions are unknown a priori, (ii) collecting evaluations is time expensive, (iii) stopping the experiments due to the controller being unstable or unsafe implies restarting the platform, which is also time consuming, and (iv) the noise is not synthetically generated, but propagated from the sensors to the cost observation we obtain. Such platform has been used as a testbench for controller learning~\citep{marco16tuning}. The experiments presented in this section build on top of those from~\citep{marco16tuning}, in which the parameters of a Linear Quadratic Regulator (LQR) controller were automatically tuned in order to improve the balancing performance. Therein, an important caveat of the learning setup was committing to a fixed heuristic cost $f_\text{heur}$, prior to starting the learning experiments. Such fixed $f_\text{heur}$ is a penalty value chosen \emph{ad hoc} for unstable controllers. As detailed in~\citep{marco16tuning}, during the learning run, a stable controller (with poor performance) that yielded a higher cost than $f_\text{heur}$ was found. Because a stable controller should not
be penalized more than an unstable one, we stopped the learning run, selected a higher $f_\text{heur}$, and restarted the learning from the beginning. Since the choice of the threshold $f_\text{heur}$ affects the decision-making of the Bayesian optimizer used therein, the data acquired with the older $f_\text{heur}$ had to be discarded for fairness. The clear practical advantage of \gpcr~is that it can overcome such situation by adapting $\thresopt$ to the worst observed stable cost, maintaining the probability mass of unstable areas always above it, and avoiding the need of restarting the learning run and discarding the data.

Additionally, in~\citep{marco16tuning} it is reported that there were two main situations that lead to instability, and, in consequence, to premature experiment detention: (i) the endeffector position leaving a safety region and (ii) the endeffector acceleration reaching a maximum value. In the context of this work, those restrictions can be seen as two external safety constraints that determine the shape of $\domsta$.

In this section, we show the advantages of using \gpcr~to take into account such constraints, and bypass the need of defining a heuristic cost. For this, we show three different tuning scenarios: (i) a 2D tuning problem with two binary constraints, (ii) idem, with one binary constraint, and one level-set constraint, and (iii) a 5D tuning problem with two binary constraints.

\subsubsection{Controller tuning with unknown constraints}
In~\citep{marco16tuning}, the design parameters of the well known Linear Quadratic Regulator (LQR)~\citep[Sec. 2.4]{anderson2007optimal} are learned from data. Therein, the performance of a balancing experiment over a finite time horizon $T$ is quantified with a quadratic cost $f(x) = \dfrac{1}{T}\left[ \sum_{t=0}^{T-1}s_t^\intercal Qs_t + u_t^\intercal Ru_t \right]$, where the state and control input trajectories are collected during an balancing experiment with control parameters $x$. Such parameters enter the diagonal of the control design weights $W_s(x)$ and $W_u(x)$, used to compute the feedback gain matrix $F(x)$ that characterizes the control strategy $u_t=F(x)s_t$. This LQR tuning problem is a particular case of \cref{ssec:controlprob}, which holds in general for a wider range of control strategies. The \gpcr~model is agnostic to the control strategy used, and thus it is also applicable to the LQR tuning. For a detailed explanation of automatic LQR tuning, we refer the reader to~\citep{marco16tuning}.

Contrary to~\citep{marco16tuning}, we consider herein a constrained tuning problem. The first constraint $g_1(x) = \max \left\lbrace s^\text{endeff}_t \right\rbrace_{t=0}^{T}$ is the maximum distance reached by the endeffector in one experiment, relative to the starting position, with constraint satisfaction determined as $g_1(x)<c_1$, where $c_1=\SI{13}{\centi\meter}$. The second constraint $g_2(x) = \max\left\lbrace \ddot{s}^\text{endeff}_t \right\rbrace_{t=0}^{T}$ is the maximum acceleration reached by the endeffector position $s^\text{endeff}_t$, with constraint satisfaction determined as $g_2(x)<c_2$, where $c_2=\SI{3}{\meter\per\second\squared}$.
Although the thresholds $c_1$ and $c_2$ are known to us, they are unknown to the model, which refines the estimates $\thresopt_1$, $\thresopt_2$ over iterations. Generally speaking, such safety thresholds are commonly deployed in robotic setups, either by the manufacturer, or by the end user, to prevent the robot from damage. While in this case we know $c_1$ and $c_2$, it is often the case that the manufacturer does not reveal such information to the end user, which motivates the need of using \gpcr~to learn such unknown thresholds.

\subsubsection{Implementation choices}
We characterize the objective cost with $Q = \diag{1,6,1,1}$ and $R=0.25$. In the 2D learning experiments, the design weights are normalized within the domain $\dom = \left[0,1\right]^D$ and parametrized as $W_s(x) = \diag{1,10^{4x_2},1,1},\; W_u(x) = 10^{4x_2-2}$, and in the 5D experiments as $W_s(x) = \diag{10^{4x_1},10^{4x_2},10^{4x_3},10^{4x_4}},\; W_u(x) = 10^{4x_5-2}$, with $x_i \in \left[0,1\right]$. In all experiments, the first point is chosen randomly in the input space. The number of BO iterations is fixed a priori. The user confidence is determined by $\delta=0.05$. In all cases we use an isometric $3/2$ Mat\'{e}rn kernel~\citep{Rasmussen2006Gaussian}, with fixed prior variance $\nu=25$. The lengthscales are fixed to $\lambda=0.1$ in the 2D experiments and to $\lambda=0.2$ in the 5D experiments. The noise for the objective is fixed to $\varnoise=0.03^2$ for $f$ and to $\varnoise=0.01^2$ for $g_1$. After completing a learning run, we test the performance of the best guess for the global minimum $\xbg$ 
on the real robot by averaging over 5 experiments, and obtaining $\meanbg$ and $\varbg$.

\subsubsection{2D controller tuning}
\begin{figure}[H]
\begin{subfigure}{0.32\linewidth}
\includegraphics[width=\linewidth]{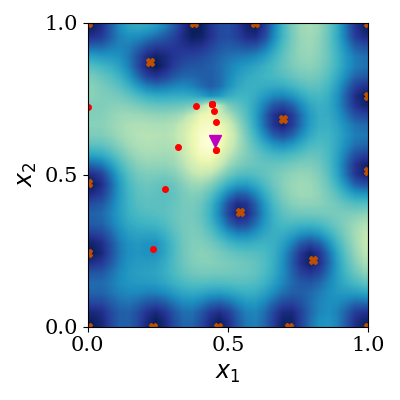}
\caption{\gpcr~mean}
\label{sfig:robot2_mean}
\end{subfigure}
\begin{subfigure}{0.32\linewidth}
\includegraphics[width=\linewidth]{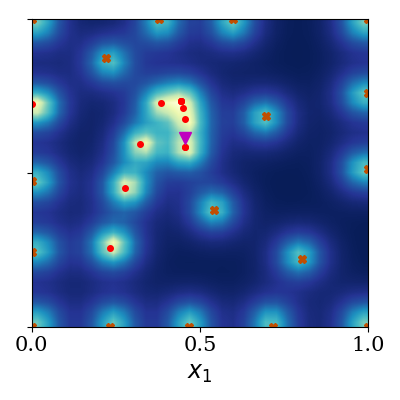}
\caption{\gpcr~std. dev.}
\label{sfig:robot2_std_dev}
\end{subfigure}
\begin{subfigure}{0.32\linewidth}
\includegraphics[width=\linewidth]{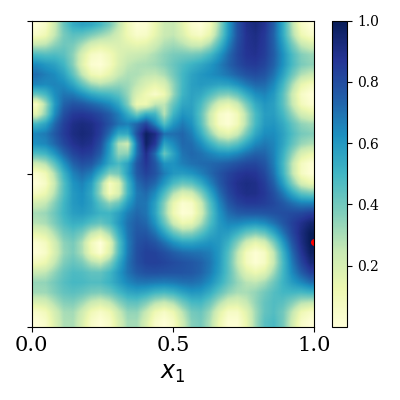}
\caption{\mesco}
\label{sfig:robot2_mesco}
\end{subfigure}
\caption{Results of the 2D controller tuning after 30 iterations using \mes~and assuming binary constraints. (a) and (b) show the predictive mean and standard deviation of \gpcr, stable evaluations (red dots), unstable evaluations (orange crosses), and the best guess estimation of the global minimum $\xbg$ (magenta triangle). }
\label{fig:robot_case2}
\end{figure}

In the first learning run, we assume $g_1$ and $g_2$ are binary constraints, and, thus, absorbed by the \gpcr~model of $f$. The hyperprior over the unstability threshold $c$ is $p(c)=\N{c}{0,10^2}$. We run our implementation of \mes~for 30 iterations, and report the results in~\cref{fig:robot_case2}. As can be seen, the best safe area appears to be found around the center of the domain. The best guess for the global minimum is found at $\xbg=\left(0.45,0.61\right)$. The performance of the global minimum is $\meanbg=0.75$, $\varbg=0.058^2$. We obtained 17 unstable evaluations and 13 stable evaluations.

\begin{figure}[h!]
\begin{subfigure}{0.24\linewidth}
\includegraphics[width=\linewidth]{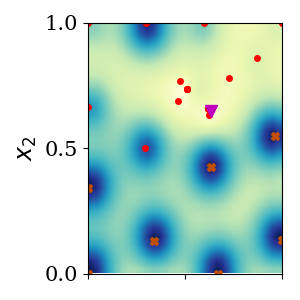}
\caption{Obj. mean}
\label{sfig:robot4_obj_mean}
\end{subfigure}
\begin{subfigure}{0.24\linewidth}
\includegraphics[width=\linewidth]{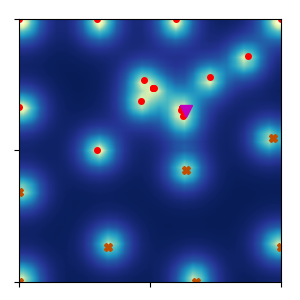}
\caption{Obj. std. dev.}
\label{sfig:robot4_obj_std_dev}
\end{subfigure}
\begin{subfigure}{0.24\linewidth}
\includegraphics[width=\linewidth]{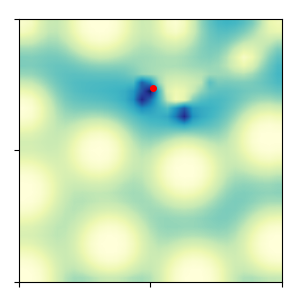}
\caption{\mesco}
\label{sfig:robot4_mesco}
\end{subfigure}
\begin{subfigure}{0.24\linewidth}
\includegraphics[width=\linewidth]{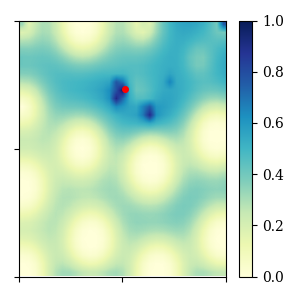}
\caption{\mesco~unconstr.}
\label{sfig:robot4_mesco_unconstr}
\end{subfigure}\\
\begin{subfigure}{0.24\linewidth}
\includegraphics[width=\linewidth]{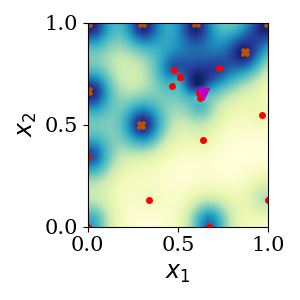}
\caption{Constr. mean}
\label{sfig:robot4_constr_mean}
\end{subfigure}
\begin{subfigure}{0.24\linewidth}
\includegraphics[width=\linewidth]{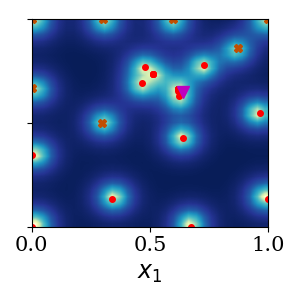}
\caption{Constr. std. dev.}
\label{sfig:robot4_constr_std_dev}
\end{subfigure}
\begin{subfigure}{0.24\linewidth}
\includegraphics[width=\linewidth]{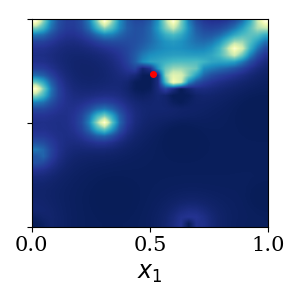}
\caption{$\Prob{g_1(x)\leq c_1}$}
\label{sfig:robot4_constr_prob}
\end{subfigure}
\caption{Results of the 2D controller tuning after 30 iterations with a level set constraint ($g_1$) and a binary constraint ($g_2$), using \mesco. 
(a) and (b) show mean and standard deviation of the objective function and (e) and (f) show the same for the constraint $g_1$.
In all 4 plots, stable evaluations (red dots), unstable evaluations (orange crosses), and the best guess estimation of the global minimum $\xbg$ (magenta triangle) are highlighted.
(c) and (d) show the acquisition function of constrained and unconstrained \mesco, respectively, and $\xnext$ (red dot).
(g) shows the probability of constraint satisfaction using the current estimate of the constraint threshold.}
\label{fig:robot_case4}
\end{figure}

In the second learning run, we assume a mixed case, in which $g_1$ is a level set-constraint and $g_2$ is a binary constraint. While $g_2$ is absorbed by the \gpcr~model for $f$, $g_1$ is modeled with another \gpcr, as shown in \cref{sfig:robot4_constr_mean,sfig:robot4_constr_std_dev}. The corresponding hyperpriors are assumed identical for both $c$ and $c_1$, with $p(c)=\N{c}{0,10^2}$.
Of special interest it is that the constraint (\cref{sfig:robot4_constr_mean}) shows unsafe evaluations in the top part of the domain, while that part appears to be stable for the objective (\cref{sfig:robot4_obj_mean}). Interestingly, in the bottom part of the domain, it happens the other way around. As a result, only the area in the center of the domain yields safe and stable controllers, which is coherent with the safe area found in the previous experiment (\cref{fig:robot_case2}). \mesco~outputs a global minimum at location $\xbg=\left(0.64,0.65\right)$, with $\meanbg=0.41$, $\varbg=0.063^2$. Evaluations to the objective yielded 23 stable and 7 unstable controllers. Evaluations on the constraint yielded (by coincidence) the same numbers. The threshold of the constraint is estimated at $\hat{c}_1=\SI{12.75}{\centi\meter}$, while the true threshold is $c_1=\SI{13}{\centi\meter}$.

\subsubsection{5D controller tuning}
\begin{figure}[h!]
\centering
\includegraphics[width=0.85\linewidth]{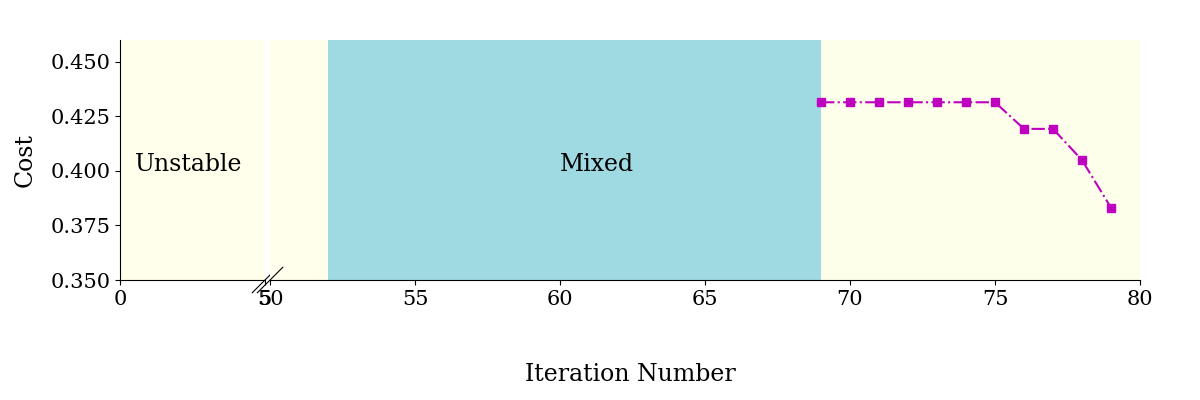}
\caption{Results of the 5D controller tuning}
\label{fig:Robot5D}
\end{figure}

Herein, we assume again that the two binary constraints are absorbed in the \gpcr~model for $f$. The best guess for the global minimum is found at $\xbg=\left( 0.51, 1.00, 0.52, 0.47, 1.00 \right)$, with $\meanbg=0.34$, $\varbg=0.045^2$. As shown in \cref{fig:Robot5D}, we obtained 65 unstable evaluations and 15 stable evaluations. The performance of the best guess is better than the one of the best guess found in the other 2D cases. At iteration 52, the first stable evaluation took place, with a cost $y_\text{52}=0.569$, which changed the value of the estimated threshold to $\hat{c}=0.602$. At iteration 69, a marginally stable controller yielded $y_\text{69}=51.64$. However, the \gpcr~model handled this situation gracefully with a new threshold estimate $\hat{c}=51.52$. With this, the probability of unstable regions is pushed above the threshold, which resolves the need of reseting the learning experiments explained in \cref{sec:robexp}, and also the major caveat observed in the approach from \citep{marco16tuning}.

\subsection{Discussion}
We have reported results on the usage of \gpcr~and \mesco~in different 2D synthetic benchmarks, and also in 2D and 5D experiments on a real robotic scenario. In summary, the benefits of \gpcr~are (i) its flexibility to learn online the instability threshold, which bypasses the need of defining a priori a heuristic penalty, and also avoids the risk of having to restart the learning experiments, (ii) its adaptability to learn online the constraint threshold whenever this is not available, and (iii) its capability of absorbing binary constraints in the model for the objective, which reduces the number of needed models, and thus, the number of hyperparameters. In addition to this, our implementation of \mesco~has shown successful results in both, synthetic benchmarks and a real robot application.\\

\noindent In the following, we discuss about some minor issues found during the experiments, as well as implementation choices.

Instead of learning the hyperparameters, we fix them to conservative values in order to ensure a wide coverage. The reason is that for such tuning scenarios, where gathering data is prohibitive, there is a high risk of overfitting to the data observed during an early stage of BO. This would yield long lengthscales and small variance, which would make the acquisition function completely myopic to other potentially interesting regions, targeting the current area as the only interesting one. On the other hand, fixing the hyperparameters yields a slow search, but the optimizer has a wider coverage over the domain, which ensures that interesting regions will not be missed.

In the 2D examples the global minimum is estimated at two nearby, but different locations. On one hand, this indicates that a higher number of evaluations would have been needed to find an agreement between the two results. Nonetheless, in both cases the Bayesian optimizer targets the area where stable controllers are present. Additionally, both \mes~and \mesco~are shown to consistently find the global minimum in the numerical benchmarks from~\cref{sec:statistical}.

In order to speed up the learning experiments on the robot, we have implemented a ``back-to-safety'' routine: When a new controller is detected as unstable, we automatically switch back to a \emph{safe controller}, which is a priori known to yield a stable controller with very poor performance. Such controller does not have any impact in the learning experiments, as it is never included as a data point in the exploration carried out by the Bayesian optimizer. It is just used to mitigate the interaction time with the robot. Even then, some controllers were too aggressive for the stable balancing to be recovered, and the robot had to be prematurely stopped by triggering safety mechanisms in place.

In practice, we found that a wide hyperprior $p(\thres)=\N{\thres}{0,\sigma_\thres^2}$ works well. 
However, it frequently occurs that during the initial iteration of \mes~and \mesco, only unstable evaluations are collected. This pushes the $\thresopt$ towards negative values, which can cause overall numerical instability issues when computing the approximate posterior~\cref{eq:posterior_approx}. To prevent this, we assign $\thresopt=0$ in the absence of stable controllers. With this choice, and small $\delta$, the probabilistic constraint will very unlikely be satisfied, which will force an exploratory strategy until the constraint is satisfied somewhere (cf. \cref{ssec:BOClevelset}).

In some situations, a stable evaluation is found at a very close to a stable one. In order to avoid numerical instability, we relax the evaluation noise of the stable point, so that the model can better adapt to such situation. For all the experiments presented in this section, we never encountered 
such issue, except for the random exploration shown in~\cref{sec:statistical}.

 \section{Conclusions}
\label{sec:conc}
We have proposed a novel non-parametric Bayesian model that solves a fundamental problem in robot learning: How to choose penalty values for unstable controllers. 
While normally such values are chosen \emph{ad hoc} after a pre-training phase of data collection, the proposed model infers them directly during learning. This alleviates the usual need of the aforementioned pre-training phase, normally used to tailor appropriate cost functions.

\noindent To this end, the Bayesian model handles two types of data: Discrete labels that indicate whether controllers are stable or unstable, and continuous-valued observations, only present when controllers are stable. The Bayesian model is approximated as a GP, which
classifies unsafe regions by pushing the predictive probability mass toward high costs. Because of the tight connection existing between classification and regression, we call this model \emph{Gaussian process for classified regression} (\gpcr).

We have also shown how this model helps to solve black-box constrained optimization problems using Bayesian optimization. Therein, unsafe regions are defined by those areas where the controller is unstable, together with those others where external constraints are violated. Our experiments show that \gpcr~can also be used to model black-box constraints and learns its threshold online, in case it is unknown.

In addition,
we have shown the benefits of \gpcr~on an robotic platform, where modeling the threshold and the penalty terms for the objective would otherwise be cumbersome, and would involve collecting extra experiments during a pre-training phase.
 
For conducting the experiments, we have extended Max-Value Entropy Search \citep{wang2017mes}, a recent information-theoretic Bayesian optimization method, to account for unknown black-box constraints, and have tested its effectiveness in simulations and real hardware.

In summary, \gpcr~is beneficial whenever designing penalty terms for the reward/cost function is problematic or requires a pre-training phase, for example learning to grasp objects with a robot arm. In addition, we have shown that \gpcr~is useful when the constraint threshold is unknown, for instance learning to pick-and-place with a robot arm under the presence of obstacles. Finally, \gpcr~is a powerful tool that can be used to learn robot controllers mitigating human intervention in the learning loop.

 \section*{Acknowledgement}
We thank Manuel W\"{u}thrich for his feedback and for the interesting discussions. We thank Felix Grimminger for the design of the pole and for his support on the hardware. We thank Vincent Berenz for his support with the software infrastructure and Julian Viereck for a key contribution to the robot code. This work was supported in part by the Max Planck Society, the Cyber Valley Initiative, the IMPRS-IS, and the German Research Foundation (DFG, grant TR 1433/1-1). Philipp Hennig gratefully acknowledges support through ERC StG Action 757275 / PANAMA. \appendix

\section{Product of two multivariate Gaussians of different dimensionality}
\label{app:prodGauss}
The product $p$ of two multivariate Gaussians of dimensions $N_1$ and $N_2$, with $N_2 > N_1$ yields another Gaussian with dimensions $N_2$. We prove this for the case in which the second Gaussian is zero-mean. To prove it, we resort to the completion of squares in the vectorial case.
\begin{equation}
p = \mathcal{N}(\Ys|\fs,\varnoise \bm{I}) 
\mathcal{N}
\left(
\begin{bmatrix} \fs \\ \fu \end{bmatrix}
\Big\vert
\begin{bmatrix} \bm{0} \\ \bm{0} \end{bmatrix}
,
\begin{bmatrix} K_{ss} & K_{su} \\ K_{us} & K_{uu} \\ \end{bmatrix}
\right)
=
a_0\exp\left\lbrace -\dfrac{1}{2}a_1\right\rbrace,
\label{eq:twoGauss}
\end{equation}
where $a_0 = \left[ (2\pi)^{\Ns} (2\pi)^{\Ns+\Nu} \deter{\varnoise\Id} \deter{K} \right]^{-1/2}$
and
\begin{equation*}
a_1 = \left[\Ys-\fs\right]^\top \stdnoise^{-2}\Id \left[\Ys-\fs\right]
+
\begin{bmatrix} \fs \\ \fu \end{bmatrix}^\top
\underbrace{
\begin{bmatrix} \bm{V}_{11} & \bm{V}_{12} \\ \bm{V}_{21} & \bm{V}_{22} \\ \end{bmatrix}
}_{K^{-1}}
\begin{bmatrix} \fs \\ \fu \end{bmatrix}. \label{eq:C}
\end{equation*}
We can reorder the terms to obtain a quadratic expresion
\begin{equation}
\begin{split}
a_1 & = 
\fs^\top \left( \bm{V}_{11} + \stdnoise^{-2} \right)\fs -2\stdnoise^{2}\Ys^\top\fs + \fs^\top \bm{V}_{12}\fu + \fu^\top \bm{V}_{21}\fs + \fu^\top \bm{V}_{22}\fu + \stdnoise^{-2}\Ys^\top\Ys  \\
& =
\begin{bmatrix} \fs \\ \fu \end{bmatrix}^\top
\underbrace{
\begin{bmatrix} \bm{V}_{11} + \stdnoise^{-2}\Id& \bm{V}_{12} \\ \bm{V}_{21} & \bm{V}_{22} \\ \end{bmatrix}
}_{C}
\underbrace{
\begin{bmatrix} \fs \\ \fu \end{bmatrix}
}_{\f}
+
\underbrace{
\begin{bmatrix} -2\stdnoise^{-2}\Ys^\top & \bm{0}^\top \end{bmatrix}
}_{\bm{b}^\top}
\begin{bmatrix} \fs \\ \fu \end{bmatrix}
+
\underbrace{
\stdnoise^{-2}\Ys^\top\Ys
}_{d} \\
& = \f^\top C\f  + \bm{b}^\top\f  + d,
. \label{eq:A_devep}
\end{split}
\end{equation}
which can be more conveniently written using the square completion in matrix form\footnote{Square completion in matrix form \citep{petersen2008matrix}: $\f^\top C\f  + \bm{b}^\top\f = \left( \f  - C^{-1}\bm{b} \right)^\top C\left( \f  - C^{-1}\bm{b} \right) + \dfrac{1}{4}\bm{b}^\top C^{-1}\bm{b}$.}
\begin{equation}
a_1 = \left[\f - C^{-1}\bm{b}\right]^\top C \left[\f - C^{-1}\bm{b}\right] + \dfrac{1}{4}\bm{b}^\top C^{-1}\bm{b} + \stdnoise^{-2}\Ys^\top\Ys.
\label{eq:completion}
\end{equation}
Using this result in~\cref{eq:twoGauss}, and defining mean $\tilde{\bm{m}} = C^{-1}\bm{b}$ and variance $\tilde{\bm{\Sigma}} = C^{-1}$, we can rewrite the product of two Gaussians as unnormalized Gaussian
\begin{align}
p & = a\cdot \exp \left\lbrace -\dfrac{1}{2} \left[\f - C^{-1}\bm{b}\right]^\top C \left[\f - C^{-1}\bm{b}\right] \right\rbrace 
\cdot
\exp \left\lbrace -\dfrac{1}{2} \left( \dfrac{1}{4}\bm{b}^\top C^{-1}\bm{b} + \stdnoise^{-2}\Ys^\top\Ys \right) \right\rbrace \nonumber \\
& = a \left[ (2\pi)^{\Ns+\Nu} \deter{{\tilde{\bm{\Sigma}}}} \right]^{1/2} \N{\f}{\tilde{\bm{m}},\tilde{\bm{\Sigma}}}
\cdot
\exp \left\lbrace -\dfrac{1}{2} \left( \dfrac{1}{4}\bm{b}^\top C^{-1}\bm{b} + \stdnoise^{-2}\Ys^\top\Ys \right) \right\rbrace \nonumber \\
& = \alpha \N{\f}{\tilde{\bm{m}},\tilde{\bm{\Sigma}}}, \nonumber
\end{align}
where the constant $\alpha$ can be simplified as

\begin{align}
\alpha & = a \left[ (2\pi)^{\Ns+\Nu} \deter{{\tilde{\bm{\Sigma}}}} \right]^{1/2} \exp \left\lbrace -\dfrac{1}{2} \left( \dfrac{1}{4}\bm{b}^\top C^{-1}\bm{b} + \stdnoise^{-2}\Ys^\top\Ys \right) \right\rbrace \nonumber \\
& = \left[ (2\pi)^{\Ns} \deter{{\tilde{\bm{\Sigma}}^{-1}}} \deter{K} \deter{\varnoise\Id} \right]^{-1/2} \exp \left\lbrace -\dfrac{1}{2} \left( \dfrac{1}{4}\bm{b}^\top C^{-1}\bm{b} + \stdnoise^{-2}\Ys^\top\Ys \right) \right\rbrace
\nonumber \\
& = \left[ e^{v} (2\pi\varnoise)^{\Ns} \deter{\tilde{\bm{\Sigma}}^{-1}} \deter{K} \right]^{-1/2}, \nonumber \\
\end{align}
with $v = \dfrac{1}{4}\bm{b}^\top K\bm{b} + \stdnoise^{-2}\Ys^\top\Ys$, and $\tilde{\bm{\Sigma}}^{-1} = K^{-1} + \begin{bmatrix} \sigma^{-2}\bm{I} & \bm{0} \\ \bm{0} & \bm{0} \\ \end{bmatrix}$.

\section{Analysis of the predictive distribution $\prob{f_*|\sobs,X,x_*}$}
\label{app:exact_pred}
The integral in the right-hand side of~\cref{eq:pred} can yield different results, depending on the assumptions we make. In~\cref{ssec:predictive}, we have followed~\cite[Sec. 3.4.2]{Rasmussen2006Gaussian} and approximated the posterior as Gaussian, which makes it possible to approximate the integral from~\cref{eq:pred} also as a Gaussian distribution $q(f_*)$, with moments~\cref{eq:pred_q_moments}. Whereas such approximation yields a tractable \gpcr~model, and works well in practice, it is also interesting to analyze how close this approximation is to the exact distribution $\prob{f_*|\sobs,X,x_*}$. For this, one would need to have the exact solution to $\prob{f_*|\sobs,X,x_*}$, which is not possible. Instead, we compare $q(f_*)$ with a better approximation $p(f_*)$ to $\prob{f_*|\sobs,X,x_*}$, obtained by relaxing some of 
the assumptions made in~\cref{ssec:predictive}, and solving the integral from~\cref{eq:pred} in a different way. The resulting distribution $p(f_*)$ is closer to the true distribution $\prob{f_*|\sobs,X,x_*}$ than $q(f_*)$, but also not exact. In addition to this, it is computationally more demanding, which makes it impractical for Bayesian optimization. To simplify notation, we refer to the predictive probability $\prob{f_*|\sobs,X,x_*}$~\cref{eq:pred} as $\prob{f_*}$. 

\begin{figure}[h!]
\centering
\includegraphics[width=0.80\columnwidth]{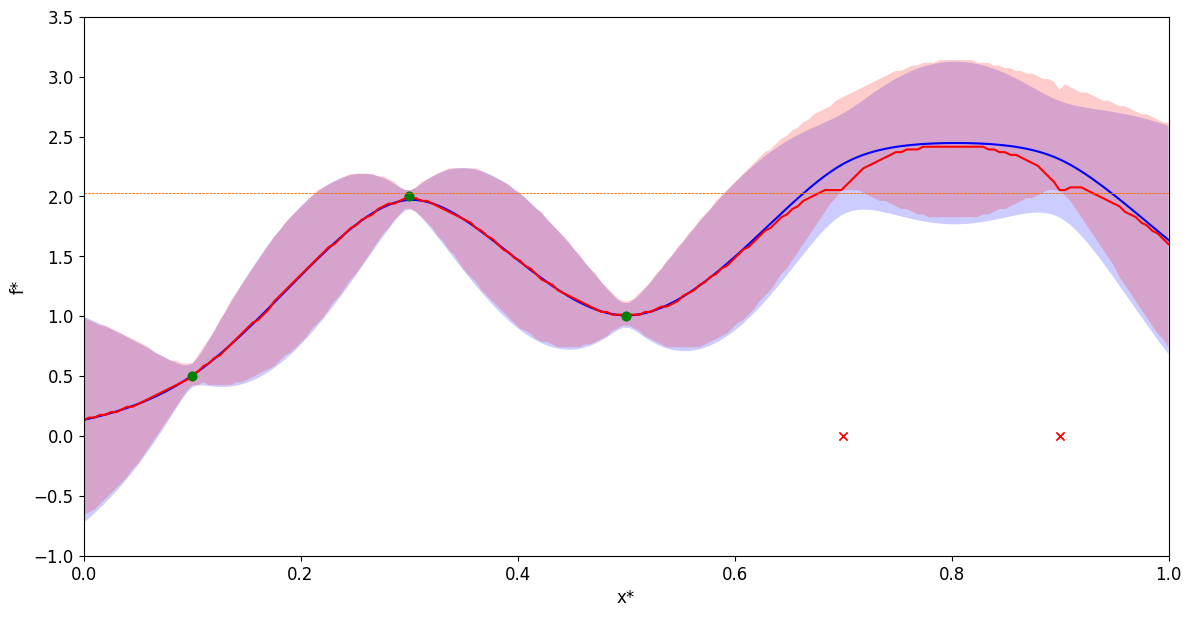}
\caption{
(red) True predictive distribution $p(f_*)$ of the Classified Regression model, conditioned on data from~\cref{ex:onedimGP}. We show the mode (red line) and the $95\%$ confidence interval (red surface) at each $x_*$ inside a discretization of the unit interval. Stable evaluations (green dots) and unstable evaluations (red crosses) are included, together with the optimal threshold $\thresopt = 2.03$ (dashed orange line). The unstable evaluations (red crosses) have been represented at $f_*=0$ for convenience.
(blue) Gaussian approximation $q(f_*)$ to the predictive distribution~\cref{eq:pred} using the \gpcr~model. We show the mean (blue line) and the $95\%$ confidence interval (blue surface).}
\label{fig:GPCR_predGP_both}
\end{figure}

\noindent We depart from $\prob{f_*}$, and use the extended prior~\cref{eq:prior} and the likelihood terms presented in~\cref{eq:post_elabo}
\begin{equation}
\begin{split}
& \prob{f_*} \propto \\
& \underbrace{\int_{\thres}^{+\infty} \dots \int_{\thres}^{+\infty}}_{\text{Unstable}} \underbrace{\int_{-\infty}^{\thres} \dots \int_{-\infty}^{\thres}}_{\text{Stable}}
\mathcal{N}(\Ys|\fs,\varnoise \bm{I})
\N{
\begin{bmatrix} \fs \\ \fu \\ f_*\end{bmatrix}
}{
\begin{bmatrix} \bm{0} \\ \bm{0} \\ 0 \end{bmatrix}
,
\begin{bmatrix} K_{ss} & K_{su} & K_{s*} \\ K_{us} & K_{uu} & K_{u*}\\ K_{*s} & K_{*u} & K_{**} \end{bmatrix}
}\dfs\dfu,
\end{split}
\label{eq:post_elabo_two}
\end{equation}
where the Heaviside functions from~\cref{eq:post_elabo} have been absorbed in the integration limits. The product of two Gaussians densities gives rise to another Gaussian density, as shown in~\cref{app:prodGauss}. Applying such transformation, \cref{eq:post_elabo_two} can be rewritten as
\begin{equation}
\prob{f_*}
\propto \underbrace{\int_{\thres}^{+\infty} \dots \int_{\thres}^{+\infty}}_{\text{Unstable}} \underbrace{\int_{-\infty}^{\thres} \dots \int_{-\infty}^{\thres}}_{\text{Stable}}
\N{\begin{bmatrix} \f \\ f_* \end{bmatrix}}{\mh,\varh}\df,
\label{eq:my_pred}
\end{equation}
where $\mh=\mh(\Ys,\varnoise)$ and $\varh=\varh(\varnoise)$ are functions of the continuous-valued observations $\Ys$ and the evaluation noise $\sigma^2$. The proportionality constant obtained from the applied transformation has been omitted for simplicity.
In order to numerically solve the integral from \cref{eq:my_pred}, we first apply the rule of conditional probability to the Gaussian, i.e., we want the conditional on $\f$ after revealing $f_*$
\begin{equation}
\mathcal{N}\left(
\begin{bmatrix} \f \\ f_* \end{bmatrix}
;
\right.
\underbrace{
\begin{bmatrix}
\hat{\bm{m}}_{\f} \\ \hat{m}_{*}
\end{bmatrix}
}_{\hat{\bm{m}}}
,
\underbrace{
\begin{bmatrix}
\hat{\Sigma}_{\f\f} & \hat{\Sigma}_{\f *} \\ 
\hat{\Sigma}_{*\f} & \hat{\Sigma}_{**} \\ 
\end{bmatrix}
}_{\hat{\Sigma}}
\bigg)
=
\N{\f}{\bm{a}(f_*),\bm{B}}\N{f_*}{\hat{m}_*,\hat{\Sigma}_{**}},
\label{eq:transf}
\end{equation}
where $\bm{a}(f_*) = \hat{\bm{m}}_{\f} + \hat{\Sigma}_{\f *}\hat{\Sigma}^{-1}_{**}\left( f_* - \hat{m}_{*} \right) $ is a linear mapping on $f_*$, and $\bm{B} = \hat{\Sigma}_{\f\f} - \hat{\Sigma}_{\f *}\hat{\Sigma}^{-1}_{**}\hat{\Sigma}_{*\f}$.
Then, we rewrite~\cref{eq:my_pred} using~\cref{eq:transf} as 
\begin{align}
p(f_*) & \propto \N{f_*}{\hat{m}_*,\hat{\Sigma}_{**}}
\underbrace{\int_{\thres}^{+\infty} \dots \int_{\thres}^{+\infty}}_{\text{Unstable}} \underbrace{\int_{-\infty}^{\thres} \dots \int_{-\infty}^{\thres}}_{\text{Stable}}
\N{\f}{\bm{a}(f_*),\bm{B}}\df \label{eq:pred_nonGaussian_int} \\
& = \N{f_*}{\hat{m}_*,\hat{\Sigma}_{**}} \F{c,f_*},
\end{align}
where we have encapsulated the multivariate integral~\cref{eq:pred_nonGaussian_int} in a functional~$\F{c,f_*}$, similarly as done for~\cref{eq:c_opti_map}. We resort also here to
the method from the authors~\citep{cunningham2011gaussian}, only that EP needs to be called for each pair of values $f_*$ and $\thres$. Of course, this involves discretizing $f_*$ for each test point $x_*$, which is computationally inefficient. We illustrate $p(f_*)$ for the simple one-dimensional~\cref{ex:onedimGP} in~\cref{fig:GPCR_predGP_both}. Therein, $f_*$ has been uniformly discretized on a fixed interval $[-1.0,+3.5]$ with $200$ divisions, and $x_*$ has also been uniformly discretized in the unit interval with $200$ divisions. We use an estimated value for the threshold $\thresopt = 2.03$, computed by maximizing the marginal likelihood~\cref{eq:c_opti}.

\cref{fig:GPCR_predGP_both} shows that $p(f_*)$ pushes 95\% of the probability mass above the threshold $\thres$, near the unstable evaluations, while it stays below the threshold, near the stable ones. In areas close to the stable evaluations, $p(f_*)$ overlaps almost perfectly with $q(f_*)$. The main differences between them arise in areas close to unstable evaluations. First, we note that $p(f_*)$ becomes an asymmetric distribution in regions close to unstable evaluation points, while it remains symmetric in regions near the stable evaluations. Second, it can be seen that $q(f_*)$ incorrectly puts some non-trivial probability mass below the threshold. When modeling a constraint $g_j$ (cf. \cref{ssec:multimeas} and \cref{ssec:mixed}), and using \mesco~\cref{eq:mesco} to perform a Bayesian optimization search, this issue could be problematic: The probability of constraint satisfaction (see \cref{eq:predictive_constraint}), computed using $q(f_*)$ is overestimated as compared to using $p(f_*)$, specially in areas close to failure points. This means that regions that are completely unsafe will not be considered. However, in practice, this issue can be solved by using more conservative user-defined thresholds $\delta$ in \cref{eq:mesco}.

\noindent This analysis shows that $q(f_*)$ is a reasonable approximation to $p(f_*)$, and thus to the true distribution \cref{eq:my_pred}.

\section{Implementation details of \mesco}
\label{app:mescoalg}
Herein, we provide additional implementation details of \mesco~to those already given in \cref{sssec:mesco}.

In \cref{alg:mESCO}, the function \textsc{ConductExperiment()} returns a real evaluation from the objective and all constraints at the same location $\xnext$. Lines 3 and 4 are optional (i.e., the threshold is only to be updated when the corresponding function is modeled with \gpcr). Inside function {\sc SampleConstrainedMinimum()}, the local optimization routine (lines 24-27) terminates when the maximum number of function evaluations $R$ is reached, or when the relative precision $\kappa$ for the optimum location is small enough. In our experiments, we set $\kappa = 10^{-3}$.

The function \textsc{LocalOptimizationStep()} (line 25) is not specifically described because it is a placeholder for any local optimization routine that can handle non-linear inequality constraints. Such routine takes as input arguments the thresholds $c$, and $\{c_j\}_{j=1}^M$ and the callable functions Ve($x_r,\mathcal{I}_r$) and \{Ve($x_r,\mathcal{I}^j_r),\thresopt_j\}_{j=1}^M$, and returns the estimated location of the constrained minimum $\tilde{x}_\text{min}$. In our experiments, we use the C\texttt{++} implementation of SLSQP from the NLOPT toolbox \citep{nlopt}. Importantly, \textsc{LocalOptimizationStep()} also receives as input argument the initial location $x_0$ (line 22), which is randomly sampled within the unit domain $\dom=[0,1]^D$ every time a new sample $\fmincons^i$ is requested. Line 36 updates the virtual data set, and keeps it as a persistent variable (or internal state), used to update the covariance matrix as described in \cref{sssec:mesco}. The randomness involved in the resulting sample $\fmincons^i$ comes from the fact that both, $x_0$ (line 22) and $y$ (line 35) are random variates.

\begin{algorithm}[H]
  \caption{min-Value Entropy Search with Unknown Level-Set Constraints (\mesco)
    \label{alg:mESCO}}
  \begin{algorithmic}[1]
    \Require{$D,S,T,R,\delta,\mu_c,\sigma_c,\{\mu_{c_j},\sigma_{c_j}\}_{j=1}^M$}
    \Require{$\sobs \gets \emptyset,\{\sobscons \gets \emptyset\}_{j=1}^M$}
    \Statex
    \Function{mESCO}{}
      \For{$t \gets 1 \textrm{ to } T$}
	    	\Let{$\thresopt$}{OptimizeThreshold($\mu_c,\sigma_c,\sobs$)} \Comment{Instability threshold}
	    	\Let{$\thresopt_j$}{OptimizeThreshold($\mu_{c_j},\sigma_{c_j},\sobscons$)$\;\forall j\in \{1,\ldots,M\}$} \Comment{Constraint threshold}
	    	\Let{$\acquimesco(\cdot)$}{SampleConstrainedMinimum($\thresopt,\sobs,\{\thresopt_j,\sobscons\}_{j=1}^M,S$)}
      	\Let{$\xnext$}{$\arg\max_{x\in\dom}\acquimesco(x)$} \Comment{Obtain next point as in \cref{eq:mesco}}
      	\Let{$y,\{y_j\}_{j=1}^M$}{ConductExperiment($\xnext$)}
      	\Let{$\sobs$}{ $\left\lbrace y,\xnext \right\rbrace$ }
      	\State $\sobscons \gets \left\lbrace y_j,\xnext \right\rbrace\;\forall j\in \{1,\ldots,M\}$
      	      	\Let{$\xbg$}{$\arg\min_{x\in\dom}\; \mu(x|\sobs), \text{ s.t. } \prod_{j=1}^M \Prob{\gn(x) \leq \gmaxn \mid \sobs_j} \geq \userconf,$}\Comment{\cref{eq:min_cost_cons_boc}}
                              \EndFor
    	\Let{$y_\text{bg}$}{ConductExperiment($\xbg$)}\Comment{Optional}
      \State \Return{$\xbg,(y_\text{bg})$}
    \EndFunction

    \Statex
    \Function{OptimizeThreshold}{$\mu_c,\sigma_c,\sobs$}
	    \State \Return{$\argmax_{\thres}\; \log\F{\thres|\sobs} - 0.5(c-\mu_c)^2/\sigma_c^2$}\Comment{MAP optimization \cref{eq:c_opti_map}}
		\EndFunction  

		\Statex
		\Function{SampleConstrainedMinimum}{$\thresopt,\sobs,\{\thresopt_j,\sobscons\}_{j=1}^M,S$}
			\For{$i \gets 1 \textrm{ to } S$}
				\Let{$\mathcal{I}_1$}{$\sobs$}\Comment{Initialize virtual data set $\mathcal{I}$ for the objective}
				\Let{$\mathcal{I}^j_1$}{$\sobscons\;\forall j=\{1,\ldots,M\}$}\Comment{Initialize virtual data sets $\mathcal{I}^j$ for the constraints}
				\Let{$x_0$}{Uniform($[0,1]^D$)}\Comment{Uniformly sample the starting point}
				\Let{r}{1}
				\While{$r \leq R$ or $||x_{r+1} -x_{r} || < \kappa$}
					\Let{$x_{r+1}$}{{\sc LocalOptimizationStep}(Ve($x_r,\mathcal{I}_r$),$\thresopt$,\{Ve($x_r,\mathcal{I}^j_r),\thresopt_j\}_{j=1}^M$,$x_0$)}
					\Let{r}{r+1}
				\EndWhile
				\Let{$\tilde{x}_\text{min}$}{$x_r$}
				\Let{$\fmincons^i$}{Ve($\tilde{x}_\text{min},\mathcal{I}_r$)}
																											\EndFor
			\State \Return{$\acquimesco(\cdot) \gets [\fmincons^i]_{i=1}^S$}
		\EndFunction 

		\Statex
		\Function{Ve}{$x,\mathcal{I}_r$}\Comment{Function to perform a ``virtual evaluation''}
			\State Sample $f \sim p(f|\mathcal{I}_r,x)$\Comment{Sample from $q(f_*)$ \cref{eq:pred_q_moments}}
			\State Add noise $y = f + \varepsilon,\;\varepsilon \sim \N{\varepsilon}{0,\sigma^2}$
			\Let{$\mathcal{I}_{r+1}$}{$\mathcal{I}_{r} \cup \{y,x\}$}\Comment{Consider $\mathcal{I}_{r}$ a persistent variable inside Ve()}
			\State \Return{$y$}
		\EndFunction
  \end{algorithmic}
\end{algorithm}

\bibliography{DataBase}

\end{document}